\documentclass[letterpaper, 10 pt, journal, twoside]{IEEEtran} % Comment this line out if you need a4paper

\IEEEoverridecommandlockouts                              % This command is only needed if 
\usepackage{multicol}
\usepackage[bookmarks=true]{hyperref}
\usepackage{graphicx}
\usepackage{hyperref}
\usepackage{dblfloatfix}
\usepackage{amsmath}
\usepackage{amssymb}
\usepackage{caption}

\usepackage{subcaption}
\usepackage{xcolor}
\usepackage{booktabs}
\usepackage{multirow,array}
\usepackage{float}
\usepackage{fp}         % For floating-point arithmetic
\usepackage[l3]{csvsimple}
\let\labelindent\relax
\usepackage{enumitem}
\usepackage{siunitx}

\usepackage{pgffor}

\usepackage{cite} % conflicts with natbib
\usepackage{amsmath,amsfonts,mathtools}
\usepackage{siunitx}
\usepackage[font=small,labelfont=bf]{caption}
\usepackage{bm}
\graphicspath{ {./figures/} }

\usepackage{algorithm}
\usepackage{algpseudocode}

\DeclareMathOperator*{\argmax}{arg\,max}

\newcommand{\comparefloats}[9]{%
    \ifdim #1 pt = #2 pt % Calculate the difference
        \ifdim #6 pt = 1 pt   
            \typeout{ TESTINGGGG! #6 #2}
            \bfseries{#1}
        \else  
            #1 
        \fi 
    \else 

    \ifdim #1 pt = #3 pt
        \ifdim #7 pt = 1 pt
            \typeout{ TESTINGGGG! #7 #3}
            \bfseries{#1}
        \else
            #1 
        \fi 
    \else 
    
    \ifdim #1 pt = #4 pt
        \ifdim #8 pt = 1 pt
        \typeout{ TESTINGGGG! #8 #4}
            \bfseries{#1}
        \else
            #1 
        \fi     
    \else
    
    \ifdim #1 pt = #5 pt
        \ifdim #9 pt = 1 pt
        \typeout{ TESTINGGGG! #9 #5}
            \bfseries{#1}
        \else
            #1 
        \fi    
    \else 
        #1
    \fi
    \fi
    \fi
    \fi 
   
}

\begin{document}
% \title{\LARGE \bf EnKode: Active Learning of Unknown Flows \\ with Koopman Operators  % Initial submission
\title{EnKode: Active Learning of Unknown Flows \\ with Koopman Operators 
% EnKODE
}

\author{Alice K. Li$^{1}$, Thales C. Silva$^{1}$, and M. Ani Hsieh$^{1}$
%\thanks{\textcolor{red}{Funding}}% <-this % stops a space
\thanks{Manuscript received: June, 14, 2024; Revised September, 1, 2024; Accepted September, 29, 2024. % }%Use only for final RAL version
% \thanks{
This paper was recommended for publication by Editor-in-Chief Tamim Asfour and Editor Jens Kober upon evaluation of the Reviewers’ comments.
This work was supported by NSF IUCRC 1939132, NSF IIS 1910308, and the University of Pennsylvania’s University Research Foundation Award} %Use only for final RAL version
\thanks{$^1$Alice K. Li, Thales C. Silva, and M. Ani Hsieh are with the General Robotics Actuation Sensing and Perception (GRASP) Laboratory at the University of Pennsylvania, Philadelphia, PA USA
        {\tt\footnotesize \{alicekl,scthales,mya\}@seas.upenn.edu}}%
\thanks{Digital Object Identifier (DOI): see top of this page.}
}

% header for the final version
\maketitle
\markboth{IEEE Robotics and Automation Letters. Preprint Version. Accepted October, 2024}
{Li \MakeLowercase{\textit{et al.}}: Active Learning of Unknown Flows
with Koopman Operators} 
\pagestyle{headings}

% commenting these out for final RA-L submission as these surpress headers and footers 
% \maketitle
% \thispagestyle{empty}

%%%%%%%%%%%%%%%%%%%%%%%%%%%%%%%%%%%%%%%%%%%%%%%%%%%%%%%%%%%%%%%%%%%%%%%%%%%%%%%%
\begin{abstract}

In this letter, we address the task of adaptive sampling to model vector fields. 
When modeling environmental phenomena with a robot, gathering high resolution information can be resource intensive. 
Actively gathering data and modeling flows with the data is a more efficient alternative.
However, in such scenarios, data is often sparse and thus requires flow modeling techniques that are effective at capturing the relevant dynamical features of the flow to ensure high prediction accuracy of the resulting models. 
To accomplish this effectively, regions with high informative value must be identified.
We propose EnKode, an active sampling approach based on Koopman Operator theory and ensemble methods that can build high quality flow models and effectively estimate model uncertainty.
For modeling complex flows, EnKode provides comparable or better estimates of unsampled flow regions than Gaussian Process Regression models with hyperparameter optimization.
Additionally, our active sensing scheme provides more accurate flow estimates than comparable strategies that rely on uniform sampling. 
We evaluate EnKode using three common benchmarking systems: the Bickley Jet, Lid-Driven Cavity flow with an obstacle, and real ocean currents from the National Oceanic and Atmospheric Administration (NOAA).

% and highlight the additional physically meaningful information that comes with linear analysis of the learnt Koopman operator.

% We can summarize some more cool results hopefully. ROM for historical maps? Better than GP? Offers modeling performance similar to a GP but with interpretable eigenfunctions and variable order model. Using historical ROM as prior is possible but not demonstrated? given sparse flow observations gathered by a robot
% \textit{Index Terms}---Environmental Monitoring; Active Sensing; Information Gathering; Koopman Operator Theory
\end{abstract}

% \thales{write a sentence or two about experiments and results.}

%%%%%%%%%%%%%%%%%%%%%%%%%%%%%%%%%%%%%%%%%%%%%%%%%%%%%%%%%%%%%%%%%%%%%%%%%%%%%%%%

\section{Introduction}
\IEEEPARstart{M}{odeling} unknown fluid flow fields is fundamental for aerial and marine robotic applications including search and rescue, oil spill mitigation, and scientific discovery, to name a few \cite{dunbabin2012robots}. %, gianfelice2022real}. 
% However, obtaining accurate flow maps is challenging, as it requires a model that effectively represents the environment dynamics in both sampled and unsampled regions, as well as access to informative data.
\textcolor{black}{However, obtaining accurate flow maps is challenging because data are difficult to acquire, and the physics of flows are not yet fully understood.}
Robots are increasingly being deployed to gather data % and actively sample the  fluid 
to build better models \cite{to2021estimation, hansen2018coverage, khodayi2023physics, chang2017motion, michini2014robotic, molchanov2015active, hollinger2016learning, salam2019adaptive, folk2024learning}.
% , due to their mobility. 
As such, since obtaining data is difficult, there is an increasing need for effective flow models and active information-theoretic sampling strategies to improve flow estimation online.

% Why are existing methods lacking to solve this problem?
% Why is Koopman a better alternative?
Although there are works that use adaptive sampling to model flows online using sparse data gathered by robots \textcolor{black}{\cite{hansen2018coverage, khodayi2023physics}}, 
% albeit few, 
%\cite{to2021estimation, hansen2018coverage, chang2017motion},
\textcolor{black}{we have not yet seen works that leverage Koopman theory.} 
% Accurate modeling of fluid flows from limited observations is challenging due to the nonlinear nature of the underlying ordinary differential equation (ODE) governing the system dynamics, as observed through the trajectory of a particle. 
Koopman Operator Theory (KOT), discovered by B.O. Koopman in 1931, is a principled \textcolor{black}{and observational bias based physics-informed method \cite{karniadakis2021physics},} for estimating the dynamics of nonlinear 
systems
\cite{koopman1931hamiltonian, mezic2013analysis, budivsic2012applied}.
This is accomplished by a coordinate transform via observables that lift the nonlinear system into % a linear space.
\textcolor{black}{a space where the dynamics are linear.}
The evolution of an observable can be considered an evolution of a hypersurface \cite{bevanda2021koopman}, 
meaning the 
% global 
linearity offered by KOT allows for high quality predictions in unsampled regions.
% \thales{Therefore, it can be insightful to apply Koopman theory to the problem of modeling unknown fluid flows from sparse data.}
Therefore, 
% since accurate modeling of fluid flows from limited observations is challenging due to the nonlinear nature of the underlying ordinary differential equation (ODE) governing the system dynamics, 
it can be insightful to apply KOT to the task of modeling flows, which are nonlinear ordinary differential equations. % (ODE).
% with access to sparse data.
% It is therefore intuitive to apply Koopman theory to the problem of modeling unknown fluid flows from sparse data.

\begin{figure}[t]
    % \vspace*{-\baselineskip} 
    \centering
    \includegraphics[width=0.48\textwidth]{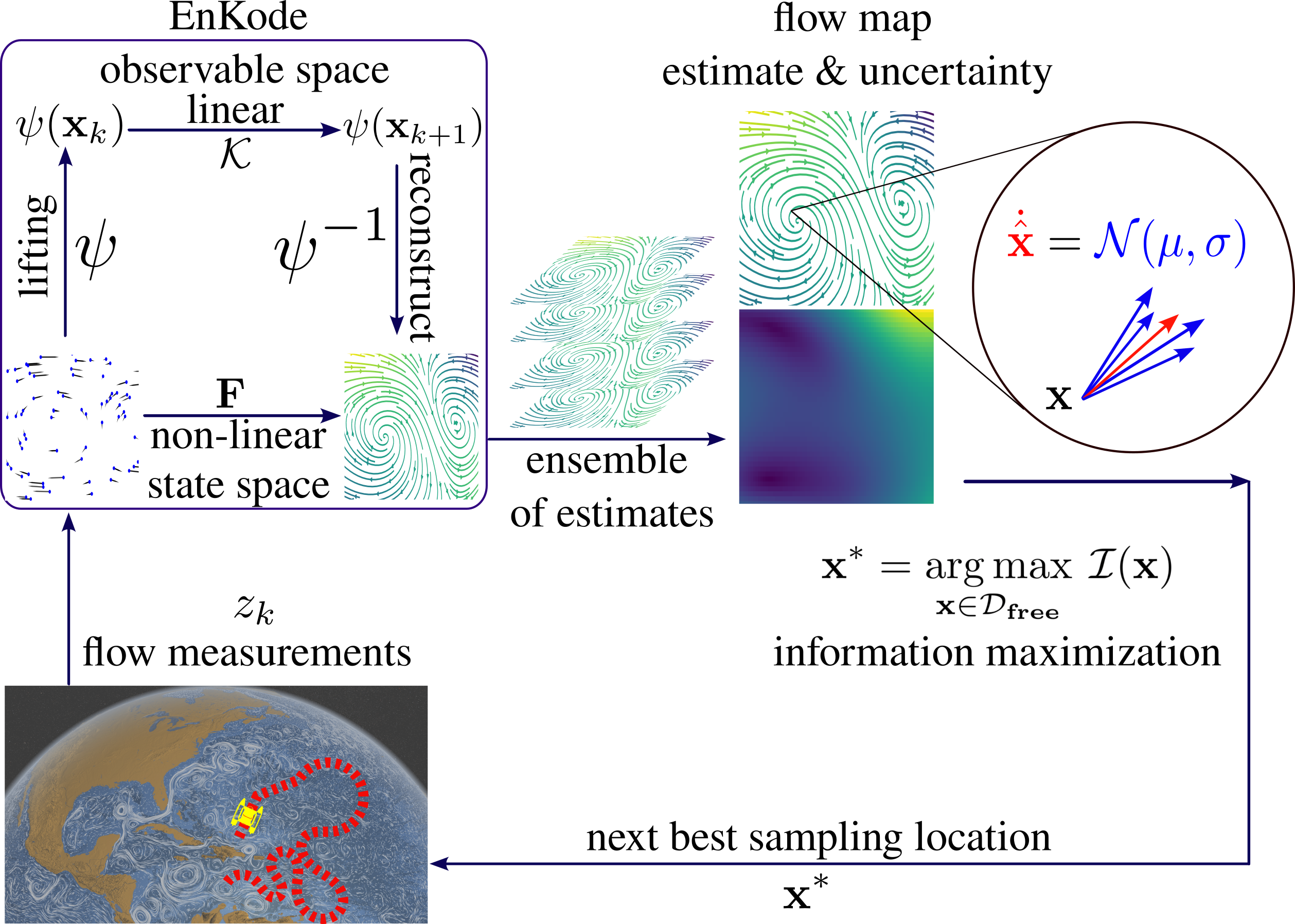}
    \caption{EnKode modeling and active sensing scheme. Flow measurements are acquired by a robot, and fed into EnKode, which combines ensemble methods and Koopman operator theory to model unknown nonlinear ordinary differential equations with model uncertainty quantification. 
    Measurements are encoded via lifting functions %into a higher dimensional and linear observable space, 
    into a space where dynamics are linear, propagated forward in time by the estimated Koopman operator, and reconstructed to produce a flow estimate and uncertainty map.
    % of both the sampled and unsampled flow regions. 
    The next best sampling location is that with the greatest epistemic uncertainty, a measure of information, providing a feedback loop for the next best sampling location.} \small
\vspace{-2mm}
\label{fig:model_uncertainties}
\end{figure}

% What are we contributing, and how does it address the above problem statement?
In this letter, we apply Koopman-based methods to enhance the efficacy of estimating flows and modeling uncertainties where data are adaptively sampled by a robot. 
This is a different problem to \cite{mezic2013analysis, otto2021koopman, williams2015data, tu2013dynamic, klus2020eigendecompositions, lusch2018deep, leask2021modal, schmid2022dynamic}, 
% which have access to large spatiotemporal datasets and focus on extracting dynamics from data. 
where the focus is on extracting dynamics from large spariotemporal datasets.
% We are particularly interested in scenarios characterized by sparse data which accurate predictions in unsampled flow regions are necessary.
Our contributions are as follows.
We present a framework, called EnKode, that is used for active modeling of unknown flows with sparse data.
%efficient?
Koopman theory is used for sample efficient modeling of flows.
We devise a novel measure of flow model uncertainty which governs the active learning strategy,
% that uses disagreement among the ensemble 
iteratively guiding a data-gathering robot towards informative sampling locations.
%\Alice{TODO: as well as Koopman modes?}
We apply our methods to various fluid flows, and compare flow modeling errors when a robot actively samples, as opposed to uniform back-and-forth coverage. We also make comparisons with the commonly used Gaussian Process regression (GP) model representation as a baseline.

% With access to trajectory data or in-situ data representing the flow velocity, the proposed method provides a model for fluid flows with no known governing equations, with a reasonable number of model parameters for online applications. The model and spectral properties of the Koopman operator are used to estimate the velocity field of the unknown flow environment, predict the trajectory of a fluid particle moving in the sparsely sampled flow, and extract the dynamics of the system. 

% Given observations of the flow, we demonstrate the effectiveness of EnKode, and make comparisons to a Gaussian Process, in the performance of modeling with a low dimensional representation, dynamics extraction, uncertainty quantification, and active learning scheme for online estimation of prototypical systems as well as real unknown ocean currents. Figure 1 summarizes the active learning loop for modeling with EnKode. And hopefully some additional summary of interesting results. 
\vspace{-2mm}
\section{Related Work}
To date, GPs \cite{rasmussen2006gaussian} have been widely used for modeling fluid flows from sparse measurements \cite{lee2019online, berlinghieri2023gaussian, gonccalves2019reconstruction, hansen2018coverage, khodayi2023physics}.
The wide adoption of GPs for estimating fluid flows is primarily due to its ease of use, the uncertainty estimation it offers, and the ability to make predictions in unsampled regions of the flow.
However, a well-known drawback of such methods is the need to pre-select a mean and covariance kernel, % based on expert knowledge. 
\textcolor{black}{where the choice governs the quality of the predictions.}
% To make high quality estimates in unsampled regions of the flow, the covariance kernel selected must represent the assumed correlation between pairs of inputs.
In addition, hyperparameters must be pre-defined. 
Despite these drawbacks, many robotics flow modeling works use GPs. 
% When using GPs, the radial basis function (RBF), also known as the squared exponential function is commonly used \Alice{cite}.
% When modeling flows, \cite{berlinghieri2023gaussian, gonccalves2019reconstruction} use the radial basis function (RBF) kernel also known as the squared exponential function, while the exponential kernel is used in \cite{hansen2018coverage}, and an incompressibility kernel is derived in \cite{lee2019online}. 

% other flow methods
Additional works have demonstrated success in modeling fluid flows in robotics using other regression techniques.
Kernel embeddings that enforce the same incompressible kernel methods derived in \cite{lee2019online} are used in \cite{to2021estimation}, combining ensemble forecasts with sensor measurements to estimate real ocean currents. 
A linear combination of RBFs are used in \cite{chang2017motion} to model ocean currents. 
In \cite{petrich2009planar}, a single singularity (either source or sink) is assumed to exist, inducing regional fluid motion, and eight flow model parameters are estimated via least squares based on measurements obtained by a robot. 

% the lack of using koopman even though it makes sense
\textcolor{black}{While these dynamics-agnostic regression methods have shown decent prediction capabilities, they are not operator-theoretic methods that model a linear functional form of dynamics.
Dynamics are estimated by lifting the system into a space in which the evolution of the system is linear.}
\textcolor{black}{Beyond prediction, this linearity enables linear analysis \cite{mezic2013analysis}, control \cite{arbabi2018data}, and extraction of physically interpretable dynamics, such as modes  \cite{mezic2013analysis} and coherent sets in\cite{salam2022}.}
\textcolor{black}{While kernel methods fit a linear model in feature space for regression, linear analysis tools and control cannot be applied, as the mapping from inputs to targets is nonlinear.}
\textcolor{black}{Therefore, using KOT, an operator-theoretic method}, to model flows with limited observations could be both more effective and favorable, but to the authors' knowledge has not been investigated yet. 
% Fluid flows are dynamical systems---the evolution of an ensemble of fluid particles can be presented as a non-linear dynamical system.
% And since the evolution is non-linear, using Koopman Operator theoretic methods become a natural choice.
% \textcolor{black}{Koopman Operator theory has been used extensively, as they can handle these non-linearities. \textbf{CITE}}

In addition to dealing with nonlinearities, the evolution of functions of the states, called observables, can be considered an evolution of hypersurfaces \cite{bevanda2021koopman}, 
meaning that the 
% global 
\textcolor{black}{linearity} offered by KOT 
allows for high quality predictions in unsampled regions. 
An example of improvements in sample efficiency is shown in \cite{han2023utility} where KOT methods were utilized for learning dexterous manipulation skills.
They observed an improvement in sample efficiency in comparison to their non-Koopman baselines.
In our work, we are interested in investigating whether Koopman-based models make high quality predictions in unsampled flow regions.

% koopman in robotics and what is most similar to ours
%\subsection{Koopman Operator Theory in Robotics}
Koopman theory has been applied to other areas in robotics, many of which involve modeling the dynamics of a robot for improved control \cite{bruder2019modeling, joglekar2023data, folkestad2020episodic, shi2021acd, abraham2019active}, rather than the dynamics of a robot's environment.
To the best of our knowledge, our work is the first to use Koopman methods for adaptive sampling of flows.
Of all these Koopman-based works listed above, \cite{abraham2019active} is the most similar in that the dynamics of quadcopters are parameterized by Koopman operators and learnt online to stabilize during free-fall. 
We are similarly interested in learning system dynamics online, because power and resource constrained robots acquiring real world data is resource intensive and can become intractable for large-scale environments \cite{li2022towards, edwards2023collaborative}.
However, to learn environmental dynamics online, prediction uncertainty must be quantified to guide robots to information rich regions. 

% existing methods for uncertainty quantification and how ours differs
%\subsection{Uncertainty Quantification for Adaptive Sampling}
Numerous techniques inspired by information theory, \textcolor{black}{such as mutual information \cite{krause2007nonmyopic}}, have been adopted to define state estimation uncertainty in robotics.
% In these estimation tasks, there is an implicit assumption that reducing the uncertainty of an estimate will reduce the estimation errors.
\textcolor{black}{Assuming that maximizing epistemic uncertainty is a proxy for maximizing information gain,} 
% when combined with active learning, 
an agent is able to query the environment and acquire more information about the state that a robot is most uncertain about \cite{ taylor2021active}.
% This is why in robotic adaptive sampling works, the next best sampling location is often based on some information theoretic measure that captures a notion of a reduction in uncertainty or information gain, based on the observed information.
% Examples of such measures include mutual information \cite{krause2007nonmyopic}, or the expected entropy reduction which is obtained by computing the difference between the trace of the prior and posterior covariance matrix provided by the GP \cite{hansen2018coverage, manjanna2018heterogeneous}. 

% An alternative approach to adaptive sampling is to define a utility function like the upper confidence bound \cite{marchant2012bayesian} that balances exploration and exploitation.

We introduce a novel method for estimating the informativeness of a sampling location when modeling flow dynamics.
Existing uncertainty estimation for KOT methods are shown in \cite{berrueta2020experimental}, where the Fisher information of each element of the Koopman operator is estimated, based on historical dynamics data gathered prior to deployment.
This is used to determine the robot control policy, providing quadcopter trajectories which maximize information gained. 
We do not assume access to prior information.
Instead, our techniques are inspired from active learning methods \cite{settles2009active} and the concept of \textit{query by disagreement} \cite{seung1992query, pathak2019self, georgakis2022learning}.
We train an ensemble of models online, and a robot samples in spatial locations corresponding to the largest disagreement in an ensemble of these models  \cite{ lakshminarayanan2017simple}.
This method does not require computations of variables in the observable space, and intuitively provides the uncertainty as the variance across the outputs of a set of models. 
Our novel Koopman-based sparse modeling, uncertainty, and adaptive sampling methods are detailed in the sections that follow.

% \Alice{Add this here.}
% In prior work \cite{salam2022} with similar model structure, model uncertainty estimates were not provided. 
% In this work, to obtain an uncertainty associated with the flow estimate for every $x \in \mathcal{D}$ in \eqref{eqn:flow_estimate}, we draw inspiration from deep ensemble models, and use multiple instances of the model to form an ensemble model, EnKode. 
% There are several variations of ensemble models \cite{lakshminarayanan2017simple, ganaie2021}. 
% In our work, we train multiple instances of the model with the same number of Fourier features.
% All models are trained on the same training data, but initialized to a different set of weights.  

\section{Problem Formulation}
% \ani{You are not taking advantage of the math notation you are laying down.}
% We consider the problem of actively mapping velocity fields of flows in a domain $\mathcal{D} \in \mathbb{R}^2$, that are \textit{a priori} unknown, based on noisy flow measurements acquired by a single mobile robot. We assume the underlying flow field that the robot is estimating is a vector valued function given by:

% Build up the maths/ingredients for understanding the problem.
\subsection{Connections Between Flows \& Koopman Operators} 
% Up to this point, we have described flow field observations as velocity vectors, as it is more intuitive for practical applications. However, from a modeling perspective, trajectory based observations, which can be obtained from velocity vectors, is the link between Koopman Operator Theoretic methods and flows. Before summarizing this connection, we first define some assumptions, defined in greater detail in Assumption 2.1 in \cite{bevanda2021koopman},
% % \Alice{revisit as this is currenty very similar, so perhaps can just cite? might help with space too}
% on the unknown fluid systems that we wish to model. 

% We consider that the fluid systems we are modeling can be represented by a continuous autonomous dynamical system of the form:
\textcolor{black}{We consider fluid systems that are assumed to be time-invariant, and can be represented by a continuous autonomous dynamical system of the form}:
\begin{equation}\label{eqn:dynamical_system}
    \mathbf{\dot{x}} = \textcolor{black}{\textbf{F}}(\mathbf{x})
    % \mathbf{x}_{k+1} = \mathbf{F}(\mathbf{x}_{k})
\end{equation}
on a smooth manifold $\mathcal{M}\in \mathbb{R}^2$, with a smooth and Lipschitz vector field $\mathbf{F} : \mathcal{M} \rightarrow \mathcal{M}$, and $\Phi(\mathbf{x}_0, k) = \mathbf{x}(k)$ is the flow of $\mathbf{F}$. 
%In our case, $\mathcal{M} \in \mathbb{R}^2$
% Given an initial condition $\mathbf{x}_0$ at time $t=0$, $\mathbf{F(x_0)} = \mathbf{x}_0 + \int_{t_0}^{t_0+t} f(\mathbf{x(\tau))}  \,d\tau$,
% where $\mathbf{F(x)}$ has a unique solution on $[0, +\infty)$  from the initial condition $\mathbf{x}$ at $t=0$.
We define $\mathcal{F}$ as the observable space, which is the 
\textcolor{black}{possibly infinite-dimensional}
space of scalar complex valued observables or observable functions $\varphi : \mathcal{M} \rightarrow \mathbb{C}$. 
% An observable is a lifting function that maps its inputs into a higher dimensional and possibly infinite scalar, complex valued space. 
\textcolor{black}{An observable is a scalar-valued lifting function measuring a relevant
property of the dynamical system.}
\textcolor{black}{It provides a coordinate transformation} %  for the system}
\textcolor{black}{into a possibly infinite-dimensional Koopman invariant space, such that the system's originally nonlinear dynamics behaves linearly}. 
% The observable must be measurable, and can be physically measurable too.}
% \textit{e.g.}, energy, position, velocity.}
% These observables are assumed to be at least continuously differentiable.
% \Alice{[Mauroy/Mezic]}.

Finally, while $\mathbf{F}$ acts on $\mathbf{x} \in \mathcal{M}$, the linear Koopman operator $\mathcal{K} : \mathcal{F} \rightarrow \mathcal{F}$ acts on observable functions $\varphi$ by composition with the evolution rule, $\mathbf{F}$. 
This is defined by:
% with the flow $\mathcal{F}$ of the vector field $\mathbf{F}$. This is defined by:
\begin{equation}\label{eqn:koopman_composition}
    \mathcal{K}\varphi = \varphi \circ \mathbf{F},
\end{equation}
where $\circ$ is the function composition. 

% Our model, 
% adaptive sampling strategy 
\textcolor{black}{EnKode}
estimates a discrete dynamics model, $\mathbf{x}_{k+1} = \mathbf{F}(\mathbf{x}_{k})$, that is sampled from the underlying process, \eqref{eqn:dynamical_system}. The discrete form of \eqref{eqn:koopman_composition} is:
% \begin{equation}\label{eqn:koopman_discrete}
    % \mathcal{K}\phi(\mathbf{x}(t_0)) = \phi(\mathbf{x}(t_0 + t)
    $ \textcolor{black}{\mathcal{K}_{\Delta t}}\varphi = \varphi(\mathbf{x}_{k+1}) $.
% \end{equation}
% Intuitively
% This tells us that when
When
\textcolor{black}{$\mathcal{K}_{\Delta t}$} acts on observable functions of the system at time $k$, the observables move along the solution curve to time $k+1$. 
% Then, using the identity function as an observable, $\phi(\mathbf{x}) = \mathbf{x}$, we obtain the flow, $\mathbf{\Phi}(\mathbf{x}(t_0))$.  % Maybe this is confusing 

% $\mathbf{F}(\mathbf{x}(t_0))$ \cite{bevanda2021koopman}. 
% Having summarized the connection between Koopman operators and flows, we can now discuss how we can use this relationship to model fluid flows. In order to obtain a $\mathcal{K}$ which is truly linear, we must find a special set of observables, called eigenfunctions. 

%\subsection{Koopman Eigenfunctions}
According to \cite{koopman1931hamiltonian, mezic2013analysis, budivsic2012applied, bevanda2021koopman}, Koopman eigenfunctions, $\psi$, are a special set of \textcolor{black}{observables such that the following holds:}
\begin{equation}\label{eqn:koopman-propagation}
    \psi(\mathbf{x}_{k+1}) = \textcolor{black}{\mathcal{K}_{\Delta t}}\psi(\mathbf{x}_k) = \lambda \psi(\mathbf{x}_k)
\end{equation}
\textcolor{black}{where $\mathbf{x}_k$ represents the state at time $k$, $\mathbf{x}_{k+1}$ is the state at time $k+1$},
%, \textit{i.e.}, trajectory-based information, 
and $\lambda$ is \textcolor{black}{an} eigenvalue of the Koopman operator. 
% Note that the first equality is another way of representing \eqref{eqn:koopman_composition}. 

With the above, we formulate our problem as follows.

\noindent \textbf{Problem 1.}
% Given a robot, noisy measurements of the flow, there are two goals: modeling + sampling 
% And the aim of the rest of the paper should be in the language of how we will solve this 
% Make sure everything I write is important
% Keep in mind that everything should be focused on how to solve the problem stated in the problem formulation 
Given a robot operating in a domain $\mathcal{D}$, \textcolor{black}{and a sampling budget $N_{total}$}, the robot must model an \textit{a priori} unknown velocity flow field \textcolor{black}{$\mathbf{F}$}: 
\begin{equation*}
\mathbf{x}_{k+1} = \textbf{F}(\mathbf{x}_{k}),
\end{equation*}
where $\mathbf{x} = [x, y]^T \in \mathcal{D}$, 
%$t$ denotes the time, 
and $\textbf{F}$
$: \mathbb{R}^2 % \times \mathbb{R}
\rightarrow \mathbb{R}^2$
based on zero-mean noisy \textcolor{black}{in-situ} flow measurements \textcolor{black}{$\mathbf{z}$} of variance $\sigma$:
\begin{equation}\label{eqn:measurements_flow}
\mathbf{z(x)} = \textcolor{black}{{\textbf{F}}}(\mathbf{x}) + \textcolor{black}{\omega },.
\end{equation}
\textcolor{black}{where $\omega \sim \mathcal{N} (0,\sigma^2)$ is an additive noise.}
\noindent \textcolor{black}{In the absence of in-situ flow measurements, $\Delta \mathbf{x} = \mathbf{F}(\mathbf{x}) \Delta t$ 
%\ref{eqn:dynamical_system} 
provides a proxy of the flow dynamics, where $\Delta \mathbf{x} = \mathbf{x}_{k+\Delta t} - \mathbf{x}_k$ is the change in robot position, between time $\Delta t$.}

% or based on position changes of a robot, $(\mathbf{x}_k, \mathbf{x}_{k+1})$ with zero-mean Gaussian noise, representing fluid flow:
%\begin{equation}\label{eqn:measurements_robot_change}
% \mathbf{z(x)} = (\mathbf{x}_k, \mathbf{x}_{k+1}) + \mathcal{N}(0,\sigma^2),
% \end{equation}

\noindent Information used to estimate a flow field are gathered iteratively by selecting the next best sampling location:
\begin{equation}\label{eqn:info_gain}
    \mathbf{x}^* = \argmax_\mathbf{x \in \mathcal{D}_{free}} \: \mathcal{I}(\mathbf{x}),
\end{equation}
where $\mathcal{I}$ is measure of informativeness of sampling from a flow field location, and $\mathcal{D}_{\mathbf{free}}$ are obstacle free and previously unsampled regions.

\section{Methodology}

% From a practical perspective, we can use \ref{eqn:koopman-propagation} to find an estimate of observables that linearize the dynamics, such that the eigenfunctions live in the span of the observables. If we then apply this coordinate transform, the system evolves linearly in the lifted space by $\mathcal{K}$.

\subsection{Estimating the Observables with EnKode}
EnKode is based on \eqref{eqn:koopman-propagation}, and finds an estimate of observables that linearize the dynamics, which is when eigenfunctions live in the span of the observables. As mentioned previously, the lifted space is possibly infinite-dimensional. However, dealing with infinite dimensions is computationally intractable. Therefore, 
like many data-driven Koopman theory works \cite{brunton2021modern, lusch2018deep, yeung2019learning, li2017extended, mauroy2019koopman}, we aim to find a finite approximation of both the eigenfunctions \textcolor{black}{$\hat{\psi}$} and the associated Koopman operator, \textcolor{black}{$\hat{\mathcal{K}}$}. 

\textbf{Training Data.}
% We assume the flow is time-invariant, similar to many works including \cite{to2021estimation, shi2021cooperative}.
\textcolor{black}{The training data of size $N$ is organized into two matrices: inputs, $\mathbf{X}_k \in \mathbb{R}^{N \times 2}$ and corresponding labels, $\mathbf{X}_{k+1} \in \mathbb{R}^{N \times 2}$.
At the start of an experiment $N=1$, \textit{i.e.,} we have a single training sample in the dataset.
The flow is then estimated; the next sampling location is determined; and a new sample is added to the dataset, making $N=2$.
This continues until the end of the experiment, where the robot is allowed a budget of $N_{total}$ samples.
For the $n$-th pair of input and label, $\mathbf{x}^n_{k} \in \mathbf{X}_k$ represents the robot's \textit{initial} position at adaptive sampling iteration $n$,
and $\mathbf{x}^n_{{k+1}} \in \mathbf{X}_{k+1}$ represents the robot's \textit{next} position, after it is evolved by the underlying flow dynamics. 
We assume the flow to be time-invariant, similar to many works including \cite{to2021estimation, shi2021cooperative}, and assume the time between any $\mathbf{x}^n_{k}$ and $\mathbf{x}^n_{{k+1}}$ pair is $\Delta t$.} 
% Data are either gathered from the change of the position of a robot 
% as in \eqref{eqn:measurements_robot_change}, 
% or from a flow sensor on-board as in %in \cite{medagoda2015autonomous}
% \eqref{eqn:measurements_flow}.
%, and integrated to give measurements of the form \eqref{eqn:measurements_robot_change}. 
\textcolor{black}{We refer to \textit{samples} as the size of the training data.}

% \textbf{Training Data.} To use Koopman Operator based methods with our in-situ measurements, observations are easily converted into be trajectory form. This is done by integrating the position of the in-situ measurement, $Z_t$ forward in time along the direction of the flow measurement, to get $Z_{t+1}$. It should be noted that changes in positions of a Lagrangian drifter or floater or tracer particles can be used, but was not done in this work as they are generally passive sensors.

% \subsubsection{Providing Structure for Observables} 
\textbf{Providing Structure for Observables.} For EnKode, $\mathbf{X}_k$
% $Z$ 
is lifted via a feature map defined by $\nu$ Fourier features. 
Each Fourier basis function is defined by a frequency \textcolor{black}{vector, $\mathbf{w}_i \in \mathbf{R}^2$}, and phase shift, $b_i$, for $i=0,...,\nu$. 
\textcolor{black}{These parameters are learnable model weights}, and are unknown prior to the modeling task and are initialized randomly.
\textcolor{black}{There are $3 \nu$ learnable parameters for EnKode.}
They are optimized online as the robot collects additional observations. 
%This feature map approximates the eigenfunctions that satisfy \eqref{eqn:koopman-propagation}. 
\textcolor{black}{The feature map $\hat{\Psi}$ is} represented as:
\textcolor{black}{
\begin{equation}\label{eqn:fourier_features}
\begin{split}
% \phi(\textbf{X}_k)= \big[\textbf{X}_k, \text{cos}(\omega_0^T \textbf{X}_k + b_0), \text{cos}(\omega_1^T \textbf{X}_k + b_1), \\
% \dots, \text{cos}(\omega_\nu^T \textbf{X}_k + b_\nu) \big].
\hat \Psi(\mathbf{X}_k)= \big[\mathbf{X}_k, \textbf{cos}( \mathbf{w}_0^T \mathbf{X}_k + b_0), \textbf{cos}(\mathbf{w}_1^T \mathbf{X}_k + b_1), \\
\dots, \textbf{cos}(\mathbf{w}_\nu^T \mathbf{X}_k + b_\nu) \big],
\end{split}
\end{equation}%
} 
\noindent where cosine is vector-valued and applied applied element-wise to its inputs.
We choose Fourier features because we empirically find that they capture the periodic and vortex-like flow dynamics well.
Additional results are shown in \cite{salam2022}. 
% Second, for more complex systems which may require additional Fourier basis functions to describe the dynamics, we may use random Fourier features instead, with small modifications to our framework. 
% Random Fourier features maps are a computationally efficient approximation of Fourier feature maps, where the features are randomly projected to a lower dimensional space. 
% See \cite{rahimi2007random} for more details.
% Additionally, EnKode offers more expressivity than a vanilla Deep Neural Network, requiring less training time.
% ; though, a NN could be used to parameterize the lifting and reconstruction with the same structure, which should provide the same results.
% Yes with Lusch et al. with their nature paper - I will leave this statement out for clarity.

% \subsubsection{Loss Function} 
\textbf{Loss Function.}
Based on this definition of the feature map, we can then find \textcolor{black}{an estimate of the Koopman operator, $\hat{\mathcal{K}}$}, using its relationship with \textcolor{black}{$\mathcal{K}_{\Delta t}$}, in \eqref{eqn:koopman-propagation}. 
We jointly learn both the feature map $\hat{\Psi}$ and the linear operator, $\hat{\mathcal{K}}$, via stochastic gradient decent, \textcolor{black}{using PyTorch, trained with the Adam optimizer and autograd}, where we minimize the following loss function over $\theta = \{ \omega, b, \textcolor{black}{\hat{\mathcal{K}}} \}$:
% \vspace{-1mm}
\begin{equation}\label{eqn:loss}
% \vspace{-1mm}
\min_{\theta} \| \textcolor{black}{\hat{\Psi}}(\mathbf{X}_{k+1}) - \textcolor{black}{\hat{\mathcal{K}}} \big( \textcolor{black}{\hat{\Psi}}(\mathbf{X}_k) \big) \|_2. 
% + \Lambda \| Z_{k} - \phi^{-1}\big(\phi(\hat{Z}_{t+1}) \big)\|. - No longer included.
\end{equation}%•
% Norms to compute loss, \textcolor{black}{$\mathcal{L}$}, are the mean squared error (MSE). 
\textcolor{black}{The mean squared error (MSE) is used to compute the loss, $\mathcal{L}$.}
This loss term is to ensure linearity of $\textcolor{black}{\hat{\mathcal{K}}}$ applied to all the lifted  measurements. \textcolor{black}{To prevent over-fitting, we apply L2
regularization on $\textcolor{black}{\hat{\mathcal{K}}}$, and weights $\theta$ via PyTorch's Adam weight decay.} 
% Weights are updated via stochastic gradient decent. % repeated.

% The first loss term is to ensure linearity of $\mathcal{K}$, and the second loss is a reconstruction loss, for lifted measurements to be `de-lifted' to the original space via $\phi^{-1}$, via a linear reconstruction with least squares. We observe that the first loss is important for low data regimes, while the second is important in high data regimes. We note that  $\phi^{-1}$ can alternatively be learnt with the same structure as \ref{eqn:fourier_features}. Works such as [X] discuss the trade-offs between either method. Additionally, to prevent overfitting, we also apply regularization on $\mathcal{K}$, and weights $\theta$. $\Lambda$ is a weighting term for the two losses--$\Lambda=1$ works well for our case. Weights are updated via stochastic gradient decent. \Alice{Something about the second loss means that it's not just Koopman based, and so the flow estimated may not satisfy the assumption above, and that the average of multiple models may also not satisfy any assumptions}

Note that in \eqref{eqn:fourier_features}, the
measurements 
% state itself 
are appended to the set of Fourier basis. This is done for two reasons. 
First, it prevents model weights from approaching zero, as a feature map with weights of zero magnitude satisfies the loss \eqref{eqn:loss}. 
Also, the system state at the next timestep, can then be found with the identity observable $\psi(\mathbf{X}_{k}) = \mathbf{X}_{k}$, without relying on any mapping for reconstruction, similar to \cite{li2017extended}. 

% \subsection{Global Linearity Provides Sampling Efficiency}
% If good approximations of the eigenfunctions are found, the system should be globally linear in the lifted space. 
% This can be understood when considering the eigenfunction evolution as the evolution of a \textit{hypersurface} which includes functions of all the initial conditions in the domain, as illustrated in \cite{bevanda2021koopman}. 
% We use this concept to demonstrate that the combination of sufficiently expressive observables offered by the Fourier features; training data that captures important system dynamics provided by the active sensing; and the global linearity offered by Koopman theory allows us to make high quality predictions in unsampled regions, and better performance compared to uniform sampling and modeling with a GP.

% \subsection{Koopman Modes}\textcolor{black}{incomplete}
% Modes for interpretability, ROM, and sampling strategy.
% Then, using \ref{eqn:modes}, we can decompose the system into dominant modes to understand the underlying physical dynamics, or can compactly represent the system with a truncated set of modes. 

% Spectral Properties?
% Being linear, is natural to consider the spectral properties of $\mathcal{K}$, which we go into more detail in the next section.

% This eigendecomposition allows us to then compute the system dynamics as the superposition of modes:

% \begin{equation}\label{eqn:modes}
%     \psi(Z_{t+1}) = \mathcal{K}\psi(Z_t) = \lambda \psi(Z_t)
% \end{equation}

\subsection{EnKode Flow Prediction and Uncertainty Estimation}
Using the 
% global 
\textcolor{black}{linearity} offered by Koopman Operators, the flow field in both sampled and unsampled regions of the flow is estimated by EnKode, and defined by the learnt lifting functions and Koopman Operator from \eqref{eqn:koopman-propagation}. 
Let us define the model output by EnKode as a flow field for $\mathbf{x} \in \mathcal{D}$: 
% \vspace{-1mm}
\begin{equation}\label{eqn:flow_estimate}
%\mathbf{\dot{\hat{x}}} = \mathbf{\hat{F}(x)} = \phi^{-1}(\mathcal{K}\phi(\mathbf{x})) \simeq \textcolor{black}{\text{approx}}.
% \vspace{-1mm}
\hat{\mathbf{x}}_{k+1} = \hat{\textbf{F}}(\mathbf{x}_{k})\simeq \textcolor{black}{\hat{\Psi}}^{-1}(\textcolor{black}{\hat{\mathcal{K}}}\textcolor{black}{\hat{\Psi}}(\mathbf{x}_{k})).
\end{equation}
The flow is estimated for test points \textcolor{black}{$\mathbf{X}' \in \mathcal{D}$}.
The forward propagated identity observable in \eqref{eqn:fourier_features} is used to construct the flow field. 
% A least squares estimate of $\phi^{-1}$ provides the same results.
% Note: Using identity or the least squares provides the same results.
% To do this, lifted measurements are `de-lifted' to the original space via $\phi^{-1}$, via a linear reconstruction with least squares. 
% We note that  $\phi^{-1}$ can alternatively be learnt with the same structure as \eqref{eqn:fourier_features}.
% Works such as \cite{pan2020physics, otto2019linearly} discuss the trade-offs between linear versus non-linear reconstruction. 
The resolution of the estimated velocity field is variable, and can be defined depending on the application. 
% This model is assumed to be updated on-board by the robot, based on observations collected over the duration of the mapping task. 
% $T$. 
To make flow predictions, we aggregate the output of $M$ models with the same structure, \textit{i.e.}, identical $\nu$, but model weights are initialized to non-identical sets of $\theta$, as in \cite{lakshminarayanan2017simple}, to obtain an average flow of $M$ models, where $M$ is the number of members of the ensemble.

To obtain the estimation uncertainty, we compute the variance of $M$ flow vectors \textcolor{black}{for all $ \mathbf{x}' \in \mathbf{X}'$}. 
Let \textcolor{black}{$\mathbf{V(\mathbf{x}^{'})}$} represent the set of $M$ 2D vector flow estimates \textcolor{black}{$\mathbf{v}_m(\mathbf{x^{'}})$} at a location $\mathbf{x}^{'}$.
We quantify uncertainty, $\mathcal{U}$, as the weighted sum of the variance of the L2 norm and circular variance:
\vspace{-1mm}
\begin{equation}\label{eqn:informative_weighting}
    \vspace{-1mm}
    \textcolor{black}{\mathcal{U}(\mathbf{x}^{'}) =  
    \newcommand{\Var}{\operatorname{Var}} \Var[\| \mathbf{V(x^{'})} \|_2]  + \beta \Var_\Theta [\mathbf{V}(\mathbf{x}^{'})],}
\end{equation}

\noindent where $\newcommand{\Var}{\operatorname{Var}} \Var_\Theta[\mathbf{V} (\mathbf{x}^{'})] = 1 - \bar{R}; \bar{R} = R / M$, and $R^2 = \big(\sum_j^M{cos(\theta_j})\big) ^ 2 + \big(\sum_j^M{sin(\theta_j})\big) ^ 2$, as defined by Fisher in \cite{fisher1995statistical}, $\theta$ is the angle of the vector from a universal $x$-axis, $\beta$ is a hyperparameter that weights the importance of the two forms of uncertainty in the modeling task.
Any number of ensemble models can be used---we show a sensitivity analysis based on the number of ensemble models used in Sec. \ref{sec:results}. 
Unless specified, we use $M=10$.

% To determine the model uncertainty, we perform fewer iterations compared to estimating the mean, as models reaching full convergence or insufficient convergence may not have meaningful variance. 

% \vspace{-10mm}
% for experiments.
% the more models, the more representative the estimate of uncertainty, which are higher in regions that are more difficult to model. 

\begin{algorithm}
    \caption{Active Learning Loop}\label{algo}
    \textcolor{black}{
    \begin{algorithmic}
    \Require $\mathcal{D}, \nu, M, \theta, N_{total}, \mathbf{X}_k = [\mathbf{x}^0_k], \mathbf{X}_{k+1} = [\mathbf{x}^0_{k+1}]$ %\,
    % \Ensure $y = x^n$
    % \State $y \gets 1$
    % \State $X \gets x$
    % \State $N \gets n$
    \For{$N \gets 1 \, \text{to} \, N_{total}$} 
        \For {m in M}
            \For {$(\mathbf{x}_k, \mathbf{x}_{k+1}) \, \text{in} \, (\mathbf{X}_k, \mathbf{X}_{k+1})$} 
                \State Predict, $\hat{\mathbf{x}}_{k+1} \gets $ EnKode($\mathbf{x}_k, \theta$)
                \State Compute loss, $\mathcal{L}  \gets $ MSELoss($\hat{\mathbf{x}}_{k+1}, \mathbf{x}_{k+1}$)
                \State Compute gradients, $\nabla_\theta \mathcal{L}$
                \State Update weights for $\hat{\Psi}$ and $\hat{\mathcal{K}}$
            \EndFor 
            \State Predict vector field $\hat{\mathbf{F}}_m(\mathbf{x})$
        \EndFor 
        \State Compute variance over M vector fields, $\mathcal{U}$
        \State Get next best sampling location, $\mathbf{x^*} =
        \underset{\mathbf{x \in \mathcal{D}_{free}}}{\argmax}
        \,\,\mathcal{U}(\mathbf{x})$
        \State Acquire training sample at $\mathbf{x^*}$, $ (\mathbf{x}^N_k, \mathbf{x}^N_{k+1}) $
        \State Append new inputs, $\mathbf{X}_k \gets [\mathbf{X}_k, \mathbf{x}^N_k]$
        \State Append new targets, $\mathbf{X}_{k+1} \gets [\mathbf{X}_{k+1}, \mathbf{x}^N_{k+1}]$
        % \State $N \gets N + 1$ 
    \EndFor
    \end{algorithmic}
    }
\end{algorithm}

\vspace{-4mm}
\subsection{Active Learning Loop}\label{sec:active_learning}

% The robot iterates through $3$ actions described below:
% : (1) active sensing, (2) flow model estimate, and (3) planning for the next best sensing location, after which we restart the procedure from action (1).
% step (1). 
% The rest of this section describes these modes in detail.

\textcolor{black}{Algorithm \ref{algo} describes the active learning loop:}

\textsc{Active Sensing} or learning allows the robot to query the environment from specific locations that benefit the map estimate, depending on the next best sensing location criterion. 
At the very start of the experiment,  $N=1$, % an initial set of random or uniform predefined measurement location(s) are selected. 
and an initial random sensing location is selected.

\textsc{EnKode updates} are performed by incorporating the actively sampled observations into the training data. 
Model weights are updated from the previous model weights. 
% Despite the additional observations, the model parameterization remains the same for the entirety of the task. 
% When updated, EnKode provides a flow estimate for any given $\mathbf{x} \in \mathcal{D}$, of any desired resolution. 
% For this, an ensemble of models is estimated, and an estimate mean $\mu$ and uncertainty $\sigma$ is provided for all $\mathbf{x}$. 
% The training of these models can be parallelized for reduced update times.
% If only one model is used, flow estimates can still be computed, but with uniform uncertainty across all $\mathbf{x}$.

% We additionally distinguish the EnKode estimate type by the amount of information that is used to compute the flow estimate. 
% If all modes are used: \textsc{All-Mode (AM)} or \textsc{Truncated-Mode (TM-\#)} where the \# represents the number of modes used for reconstruction.

\textsc{Planning} for the next best sampling location is determined by the concept of \textit{query by disagreement} \cite{settles2009active}. 
\textcolor{black}{For this, we assume that minimizing uncertainty is a proxy for maximizing information gain $\mathcal{I}$ in \ref{eqn:info_gain}, as in \cite{pathak2019self}.}
% When used in our scenario, we define the
The most informative sampling location is that with greatest disagreement in prediction in the $M$ models.
The most informative point $\mathbf{x^* \in \mathcal{D}_{free}}$, given model prediction uncertainty $\mathcal{U}$ is:
\vspace{-1mm}
\begin{equation}\label{eqn:informativeness}
    \vspace{-1mm}
    \textcolor{black}{\mathbf{x}^* = \argmax_\mathbf{x \in \mathcal{D}_{free}} \: \mathcal{U}(\mathbf{x}).}
\end{equation}
% \noindent \textcolor{black}{where $\mathcal{U}$ is the model prediction uncertainty. }

% an engineering detail - to include?:
% and the expected value of the norm of the random vectors \mathbb{E}[\|\mathbf{V}\|_2] is used to weigh more importance on larger flows.

% \begin{equation}
%     \mathbf{x}^* = \argmax_\mathbf{x} \mu(\mathbf{\dot{\hat{x}}}) + \beta \sigma(\mathbf{\dot{\hat{x}}}).
% \end{equation}
% where $\beta$ is a trade off term between exploration and exploitation. A high $\beta$ value encourages exploitation and biases towards regions that lack agreement, to minimize the entropy across models. With sufficient training samples, this generally corresponds to regions where features of the dynamics are finer, as we demonstrate in Section \ref{sec:results}. On the other hand, a low $\beta$ term encourages exploration, which entails sampling in regions with estimates of high flow energy.

We note the sampling strategy is myopic and does not consider distance travelled. 
This is done intentionally to remove all other independent variables, allowing us to systematically assess whether sampling in spatial locations with maximum uncertainty estimated by EnKode reduces modeling error. 
% An additional term can easily be added to sample with minimal travel cost, but is not in the scope of this work.

\begin{table*}[htp]
    % \fontsize{6}{4} 
    \vspace{6mm}
    \centering
    \caption{\textsc{Performance metrics for active sampling up to $N \leq 36$ with EnKode and GP.}} 
    % (m32 and rbf kernel) \textsc{w/ opt} on Bickley Jet, Lid-Driven Cavity Flow, and Ocean Currents. 
    % showing active sampling is preferred over uniform sampling in more complex flows.}
    \NewDocumentCommand{\B}{}{\fontseries{b}\selectfont}
    \centering
    \resizebox{\textwidth}{!}{%
    \begin{tabular}{
        @{}
        l
        c c c |
        c c c |
        c c c |
        c c c |
        c c c |
        c c c |
        @{}
    }
        \toprule
        % & \multicolumn{18}{c}{\textsc{Active Sampling}} \\ 
        % \cmidrule(lr){2-19}
        & \multicolumn{6}{c}{Bickley} & \multicolumn{6}{c}{Cavity} & \multicolumn{6}{c}{Ocean} \\
        \cmidrule(lr){2-7} \cmidrule(lr){8-13}
        \cmidrule(lr){14-19}
        & \multicolumn{3}{c}{CS ($\uparrow$)} & \multicolumn{3}{c}{ME ($\downarrow$)} & \multicolumn{3}{c}{CS ($\uparrow$)} & \multicolumn{3}{c}{ME ($\downarrow$)} & \multicolumn{3}{c}{CS ($\uparrow$)} & \multicolumn{3}{c}{ME ($\downarrow$)} \\
        % & \multicolumn{2}{c}{EnKode} & \multicolumn{2}{c}{GP-m32} & \multicolumn{2}{c}{GP-rbf}\\
        \cmidrule(lr){2-7} \cmidrule(lr){8-13}
        \cmidrule(lr){14-19}
        N & {EnK} & {GP-m32} & {GP-rbf}
        &{EnK} & {GP-m32} & {GP-rbf}
        &{EnK} & {GP-m32} & {GP-rbf}
        &{EnK} & {GP-m32} & {GP-rbf}
        &{EnK} & {GP-m32} & {GP-rbf}
        &{EnK} & {GP-m32} & {GP-rbf}\\
        \midrule
        % \csvreader[late after line=\\, head to column names,
        %   table foot=\hline,
        %   % late after line=\\\hline,
        % ]{metrics_v2.csv}{}
        % {%
       
        %     \Samp 
        %     & \comparefloats{\a}{0.256}{0.456}{0.644}{0.791}{1}{1}{0}{1}
        %     & \comparefloats{\b}{0}{0}{0}{0}{0}{0}{0}{0} 
        %     & \comparefloats{\c}{0}{0}{0}{0}{0}{0}{0}{0} 
        %     & \comparefloats{\d}{0.012}{0.010}{0.008}{0.005}{1}{1}{1}{1}
        %     & \comparefloats{\e}{0}{0}{0.008}{0}{0}{0}{1}{0}
        %     & \comparefloats{\f}{0}{0}{0}{0}{0}{0}{0}{0}
        %     & \comparefloats{\g}{0}{0}{0.668}{0}{0}{0}{1}{1}  
        %     & \comparefloats{\h}{0.608}{0}{0}{0.728}{1}{1}{0}{1}
        %     & \comparefloats{\i}{0}{0}{0}{0}{0}{0}{0}{0}
        %     & \comparefloats{\j}{0}{0.051}{0.039}{0.030}{0}{1}{1}{1}
        %     & \comparefloats{\k}{0.062}{0}{0}{0}{1}{0}{0}{0}
        %     & \comparefloats{\l}{0}{0}{0}{0}{0}{0}{0}{0}
            
        %     & \comparefloats{\m}{0}{0}{0.531}{0.639}{0}{0}{1}{1}
        %     & \comparefloats{\n}{0.440}{0}{0}{0}{1}{0}{0}{0}
        %     & \comparefloats{\o}{0}{0.474}{0}{0}{0}{1}{0}{0}
            
        %     & \comparefloats{\p}{0}{0}{0}{0}{ 0}{0}{0}{0}
        %     & \comparefloats{\q}{0}{0}{0.013}{0.011}{0}{0}{1}{1}
        %     & \comparefloats{\r}{0.017}{0}{0}{0} {1}{0}{0}{0}
        % }

        9&\textbf{0.256}&0.021&-0.037&  \textbf{0.012}&\textbf{0.012}&0.013    &0.284&\textbf{0.608}&0.323      &0.063&\textbf{0.062}&0.073      &0.234&\textbf{0.440}&0.319      &0.074&\textbf{0.017}&\textbf{0.017} \\
        16&\textbf{0.456}&0.272&0.060&  \textbf{0.010}&0.011&0.013              &0.573&\textbf{0.604}&0.601              &\textbf{0.051}&0.057&0.065      &0.385&0.354&\textbf{0.474}      &0.024&0.018&\textbf{0.016} \\
        25&\textbf{0.644}&0.566&0.148&  \textbf{0.008}&\textbf{0.008}&0.015     &\textbf{0.668}&0.649&0.636     &\textbf{0.039}&0.055&0.054      &\textbf{0.531}&0.503&0.479      &\textbf{0.013}&0.016&0.017 \\
        36&\textbf{0.791}&0.715&0.231&  \textbf{0.005}&0.007&0.015              &\textbf{0.728}&0.657&0.638              &\textbf{0.030}&0.051&0.054      &\textbf{0.639}&0.599&0.510      &\textbf{0.011} &0.013&0.016 \\ 

        % CS & 0.434 & 0.783 & 0.283 & 0.697 & 0.060 & 0.230 & 0.579 & 0.728 & 0.601 & 0.656 & 0.600 & 0.647 &  0.385 & 0.639 & 0.354 & 0.600 & 0.474 & 0.525\\
        % ME & 0.009 & 0.005 & 0.011 & 0.007 & 0.013& 0.015 &0.050&0.030 &0.056&0.050& 0.065 & 0.052 & 0.023 &0.011&0.018&0.012 & 0.016 & 0.016  \\
        \bottomrule
    \end{tabular}}
    % Active sampling with EnKode gives comparable or better performance than that with a GP with m32 or rbf kernel.
    \label{tab:metrics_active}
    \vspace{3mm}
\end{table*}

\begin{table*}[htp]
    % \fontsize{6}{4}  
    \centering 
    \caption{\textsc{Performance metrics for uniform sampling up to $N \leq 36$ with EnKode and GP.}}
    % (m32 and rbf kernel) \textsc{w/ opt} on Bickley Jet, Lid-Driven Cavity Flow, and Ocean Currents. CS values closer to $1$ are desired, while ME closer to $0$ are desired. Uniform sampling is preferred for more modeling flows dominated by uniform bulk flow \textit{e.g.} Lid-Driven Cavity.}
    \NewDocumentCommand{\B}{}{\fontseries{b}\selectfont}
    \centering
    \resizebox{\textwidth}{!}{%
    \begin{tabular}{
        @{}
        l
        c c c |
        c c c |
        c c c |
        c c c |
        c c c |
        c c c |
        @{}
    }
        \toprule
        % & \multicolumn{18}{c}{\textsc{Uniform Sampling}} \\ 
        % \cmidrule(lr){2-19}
        & \multicolumn{6}{c}{Bickley} & \multicolumn{6}{c}{Cavity} & \multicolumn{6}{c}{Ocean} \\
        \cmidrule(lr){2-7} \cmidrule(lr){8-13}
        \cmidrule(lr){14-19}
        & \multicolumn{3}{c}{CS ($\uparrow$)} & \multicolumn{3}{c}{ME ($\downarrow$)} & \multicolumn{3}{c}{CS ($\uparrow$)} & \multicolumn{3}{c}{ME ($\downarrow$)} & \multicolumn{3}{c}{CS ($\uparrow$)} & \multicolumn{3}{c}{ME ($\downarrow$)} \\
        % & \multicolumn{2}{c}{EnKode} & \multicolumn{2}{c}{GP-m32} & \multicolumn{2}{c}{GP-rbf}\\
        \cmidrule(lr){2-7} \cmidrule(lr){8-13}
        \cmidrule(lr){14-19}
        N & {EnK} & {GP-m32} & {GP-rbf}
        &{EnK} & {GP-m32} & {GP-rbf}
        &{EnK} & {GP-m32} & {GP-rbf}
        &{EnK} & {GP-m32} & {GP-rbf}
        &{EnK} & {GP-m32} & {GP-rbf}
        &{EnK} & {GP-m32} & {GP-rbf}\\
        \midrule

        9&0.017&\textbf{0.030}&-0.032&   \textbf{0.012}&\textbf{0.012}&0.013     &0.410&\textbf{0.596}&0.551                      &\textbf{0.063}&\textbf{0.063}&0.064      &0.428&\textbf{0.526}&0.513      &0.032&\textbf{0.014}&\textbf{0.014} \\
        16&\textbf{0.204}&0.185&0.119&   \textbf{0.012}&\textbf{0.012}&0.013                       &\textbf{0.633}&\textbf{0.633}&0.626    &\textbf{0.059}&\textbf{0.059}&0.063      &0.437&\textbf{0.560}&0.523      &0.021&\textbf{0.013}&0.014 \\
        25&\textbf{0.322}&0.267&0.243&   \textbf{0.009}&0.010&0.013                       &0.643&\textbf{0.658}&0.642                      &\textbf{0.042}&\textbf{0.042}&0.058      &0.498&\textbf{0.578}&0.524      &0.014&\textbf{0.013}&0.014 \\
        36&0.463&\textbf{0.471}&0.324&   0.007&\textbf{0.005}&0.013              &0.769&\textbf{0.780}&0.654             &0.025&\textbf{0.024}&0.042      &\textbf{0.621}&0.587&0.527      &\textbf{0.011}&0.013&0.013 \\
        
        % \csvreader[late after line=\\,
        % head to column names,
        %   ]{metrics_uniform_v2.csv}{}{%
        %     % \Samp & \num{\EnKa} & \num{\GPma} & \num{\GPra} & \num{\EnKb} & \num{\GPmb} & \num{\GPrb} & \num{\EnKb} & \num{\GPmc} & \num{\GPrc} & \num{\EnKc} & \num{\GPmd} & \num{\GPrd} & \num{\EnKd} & \num{\GPme} & \num{\GPre} & \num{\EnKf} & \num{\GPmf} & \num{\GPrf}
            
        %    \Samp 
        %    & \comparefloats{\a}{0}{0}{0}{0}{0}{0}{0}{0}
        %    & \comparefloats{\b}{0}{0}{0}{0}{0}{0}{0}{0}
        %    & \comparefloats{\c}{0}{0}{0}{0}{0}{0}{0}{0}
        %    & \comparefloats{\d}{0.012}{0.012}{0}{0}{0}{0}{0}{0}
        %    & \comparefloats{\e}{0}{0.012}{0}{0}{0.005}{0}{0}{0}
        %    & \comparefloats{\f}{0}{0}{0}{0}{0}{0}{0}{0}
        %    & \comparefloats{\g}{0}{0}{0}{0}{0}{0}{0}{0}
        %    & \comparefloats{\h}{0.430}{0}{0.633}{0}{0.780}{}{0}{0}
        %    & \comparefloats{\i}{0}{0}{0.633}{0}{0}{0}{0}{0}
        %    & \comparefloats{\j}{0}{0.063}{0}{0}{0}{0}{0}{0}
        %    & \comparefloats{\k}{0.063}{0}{0}{0}{0}{0.024}{0}{0}
        %    & \comparefloats{\l}{0}{0}{0}{0}{0}{0}{0}{0}
        %    & \comparefloats{\m}{0}{0}{0}{0}{0}{0}{0}{0}
        %    & \comparefloats{\n}{0}{0.526}{0}{0}{0}{0}{0}{0}
        %    & \comparefloats{\o}{0.499}{0}{0}{0}{0}{0}{0}{0}
        %    & \comparefloats{\p}{0}{0}{0}{0}{0}{0}{0}{0}
        %    & \comparefloats{\q}{0}{0}{0}{0}{0}{0}{0}{0}
        %    & \comparefloats{\r}{0}{0}{0}{0}{0}{0}{0}{0}
        % }

        \bottomrule
    \end{tabular}}
    % \caption{Performance metrics for uniform sampling up to $N \leq 36$ with EnKode and GP (m32 and rbf kernel) \textsc{w/ opt} on Bickley Jet, Lid-Driven Cavity Flow, and Ocean Currents. CS values closer to $1$ are desired, while ME closer to $0$ are desired. Uniform sampling is preferred for more modeling flows dominated by uniform bulk flow \textit{e.g.} Lid-Driven Cavity.}
    \label{tab:metrics_uniform}
    % \vspace{1mm}
\end{table*}

\section{Experiments}

%\section{Experiments}
% We conduct 250 experimental runs in total. 
 \textcolor{black}{We conduct various experiments with $N \leq 36$}. First, we use adaptive sampling to model vector fields using EnKode without hyperparameter optimization, versus GP with hyperparameter optimization \textsc{(w/ opt)}, and GP without hyperparameter optimization when number of samples $\textcolor{black}{N} < 10$ \textsc{(w/o opt)}. 
% \textcolor{red}{Hyperparameters used for EnKode are: number of iterations:, learning rate:, weight decay.}
We then make comparisons of adaptive versus uniform sampling, akin to back-and-forth patterns for information gathering commonly adopted in real-world applications. We evaluate the performance of EnKode on three different fluid flow domains: Bickley Jet, the Lid-driven Cavity flow with an obstacle, and real ocean currents from the National Oceanic and Atomspheric Administration, from ERDDAP \cite{simons2022erddap}. Finally, we assess the effect of three parameters of interest (number of Fourier features $\nu$ in \eqref{eqn:fourier_features}, number of ensemble models $M$, weighting $\beta$ defined in \eqref{eqn:informativeness}.  

% lawn mower patterns %a back-and-forth - I got told once that lawn mower wasn't a proper term, but back-and-forth/Zamboni was

%\section{Evaluation Metrics}
% \Alice{TODO: Decide. perhaps can simply integrate this in results}}

\begin{figure*}
    \centering    \includegraphics[width=1.0\textwidth]{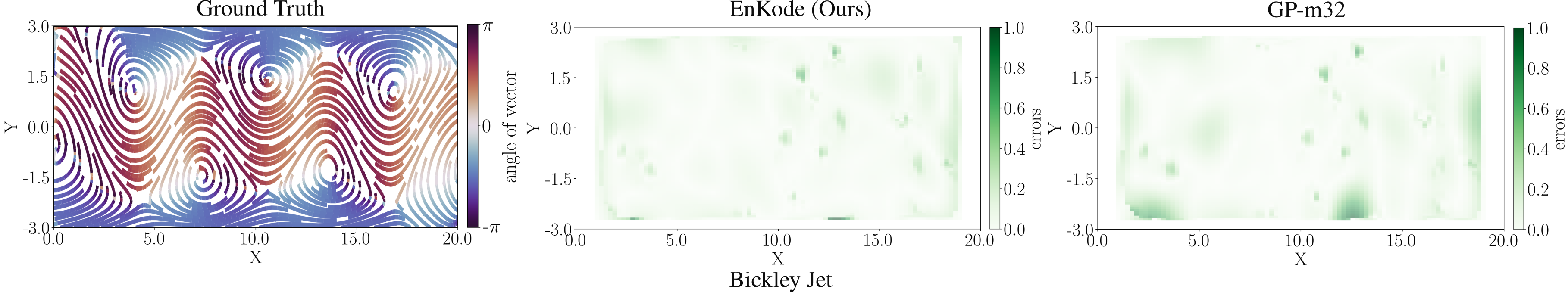}
    \caption{Pixel-wise EPE after actively modeling the flow of the Bickley Jet, for $N_{total} = 36 $. Overall EnKode error is lower than that of GP-32 \textsc{w/ opt}. There are localized regions with errors for the GP that do not appear in the EnKode estimate.}
    \normalsize
\label{fig:flows_bickley}
\end{figure*}

The following metrics are used to assess Enkode's modeling and adaptive sampling performance:
\begin{enumerate}[left=0pt]
    % \item \textit{Training Time (TT)}. The time in seconds taken to estimate the model prediction and uncertainty. 
    \item \textit{End-to-End Point Error (EPE).} We take inspiration from optical flow metrics, which apply to our flow estimation task. 
    For spatial plots, for every $\mathbf{x} \in \mathbf{X}^{'}$, we compute the Euclidean norm of point-wise EPE: $\| \mathbf{\dot{x}} - \mathbf{\dot{\hat{x}}} \|_2$ \cite{otte1994optical}.
    % ; otherwise, as a measure of error across an entire domain, we report on average EPE, $\frac{1}{n} \sum_{i=1}^{n} \| \mathbf{\dot{x}}_i - \mathbf{\dot{\hat{x}}}_i \|_2$, where $n$ is the total number of vectors in the field.
    % Wherever relevant, we also report on the variance of errors. 
    \item \textit{Cosine Similarity (CS)}. For CS, we compute point-wise errors across the domain $CS = \frac{\mathbf{\dot{x}} \cdot \mathbf{\dot{\hat{x}}}}{\| \mathbf{\dot{x}} \|_2 \| \mathbf{\dot{\hat{x}}} \|_2} \in [-1, 1]$, and report the average CS across the entire domain. %\cite{barron1994performance}. 
    The closer the CS to $1$, the more aligned the vector. A value of $-1$ signifies vectors are oriented in opposite directions.
    \item \textit{Magnitude Error (ME)}. While EPE may capture a notion of magnitude error, we additionally evaluate on ME, which removes any angular error dependence. ME is defined as $\Big\| \| \mathbf{\dot{x}} \|_2 - \| \mathbf{\dot{\hat{x}}} \|_2 \Big\|_2$.
    % \item \textit{Trajectory Root Mean Squared Error (RSME)}. The similarity between trajectories capturing the evolution of particles, whose motion is governed by the estimated and actual flow models. This is estimated for $20$ steps of propagation, and $256$ randomly selected particles. 
    % Note: Ani recommended switching this out with magnitude difference.
\end{enumerate}
% might not need these as can report on average EPE 
% \subsubsection{Mean Squared Error (MSE)}
% $\sqrt{\frac{1}{n} \sum_{i=1}^{n} ( \dot{x}_i - \hat{\dot{x}}_i)^2}$. 
% \subsubsection{Mean Absolute Error (MAE)} An L1-norm error: $\frac{1}{n} \sum_{i=1}^{n} \left| \dot{x}_i - \hat{\dot{x}}_i \right|$, that does not penalize outliers like MSE.
% \ani{We need to discuss these in detail.  I am not sure how you are doing the evaluation and why it makes sense to do the evaluation the way you did.  I would like to understand this better.}
% what are our experiments; what do we want to achieve.
Every CS, EPE, and ME curve and sensitivity results represent an aggregate of $10$ experimental trials. 
% Sensitivity results are an aggregate of $10$ experimental trials.
For the GP baseline, we use GPy \cite{gpy2014} and their L-BFGS hyperparameter optimization functionality, either performed at every sampling iteration (\textsc{w/ opt}), or at every sampling iteration \textcolor{black}{$N$} $< 10$ samples (\textsc{w/o opt}). 
We make comparisons with two commonly used kernels: (1) Radial Basis Function (GP-rbf), also known as the squared exponential function: \textcolor{black}{$k(\mathbf{x}_i, \mathbf{x}_j)=\text{exp}\Big( -\frac{d(\mathbf{x}_i, \mathbf{x}_j)^2}{2l^2}\Big)$, where $d$ denotes Euclidean distance between its inputs, and $l$ is a lengthscale,} \textcolor{black}{and (2) Mat\'ern32 (GP-m32): $\sigma_f^2\Big(1+\sqrt{\frac{3d(\mathbf{x}_i, \mathbf{x}_j)^2  }{l^2}}\Big) \text{exp}\Big(-\sqrt{\frac{3d(\mathbf{x}_i, \mathbf{x}_j)^2}{l^2}} \Big)$}. 
\textcolor{black}{The GP has $4$ learnable parameters: noise variance $\sigma _n$, signal variance $\sigma_f$,  and length-scale parameters for each dimension $l_1, l_2$}.
% Due to space limitations, we only show results for GP-m32, which performs comparably or better than RBF in all flows and for both sampling methods, at the given sample size. 
\textcolor{black}{Next, we define variables with standard GP notation, making connections with our EnKode variables wherever possible.
GP training inputs are $\mathbf{X}^{GP} = \mathbf{X}_{k}$. 
The labels are $y^{GP} = \mathbf{X}_{k+1}$.
A GP kernel $K(\mathbf{X^{'}}, \mathbf{X}^{GP})$ denotes a matrix of covariances evaluated at all pairs of test points $\mathbf{X^{'}}$ and training points $\mathbf{X}^{GP}$.
Given a grid of initial position as input, $\mathbf{X^{'}}$, the GP prediction of the next position, $\mathbf{y^{'}} = K(\mathbf{X^{'}}, \mathbf{X}^{GP})[K(\mathbf{X}^{GP}, \mathbf{X}^{GP}) + \sigma_n^2 I]^{-1}\mathbf{y}.$ 
Since \cite{verner2024efficient} demonstrates similar uncertainty quantification between ensemble GPs and standard GPs, and by construction, \textcolor{black}{standard} GPs offer prediction uncertainty, we use standard GP uncertainty to determine the next best sampling location: $\mathbf{x}^* = \argmax_{\mathbf{x} \in \mathbf{X}^{'}} \: \Sigma(\mathbf{X^{'}}, \mathbf{X^{'}})$ where
$\Sigma(\mathbf{X^{'}},\mathbf{X^{'}}) = K(\mathbf{X}^{'}, \mathbf{X}^{'}) - K(\mathbf{X}^{'}, \mathbf{X}^{GP})[K(\mathbf{X}^{GP}, \mathbf{X}^{GP}) + \sigma_n^2 I]^{-1}K(\mathbf{X}^{GP}, \mathbf{X}^{'})$.
The location corresponds to $ \mathbf{x}' \in \mathbf{X}'$ with maximum uncertainty, given by the diagonal entries of $\Sigma$.}  

% \vspace{-1mm}
\section{Results and Discussion}\label{sec:results}
\subsection{Performance Overview}
Overall, the proposed Koopman based modeling, uncertainty quantification, and adaptive sampling scheme is suitable for learning unknown flows online.
When actively modeling flows, EnKode achieves either comparable or higher CS and lower ME errors by the end of the sampling task, in comparison to the GP with or without hyperparameter optimization. 
This is demonstrated by the errors in Table \ref{tab:metrics_active}, for all flows.
EnKode errors converge more stably with increasing number of samples, compared to the GP. 
Active sampling is preferred over uniform sampling when flows are dominated by more complex structure, and less uniform flow (uni-directional), \text{e.g.}, Bickley Jet. 
This is observed when comparing Table \ref{tab:metrics_active} with Table \ref{tab:metrics_uniform} for each flow system.
In general, when restricted to very few samples (on the order of $9-16$), uniform sampling is preferable, as uniformly distributed samples capture the general flow structure better.
\textcolor{black}{Unless stated, we use $\nu = 64$, for all flow models}. Due to space limitations, we show qualitative EPE, and tabulate and plot quantitative ME and CS for varying $N$.
% in Fig. \ref{fig:metrics_bickley}, \ref{fig:metrics_cavity}, \ref{fig:metrics_ocean}, and Table \ref{tab:metrics_active}, \ref{tab:metrics_uniform}, \ref{tab:sensitivity_beta}, EPE is not shown but follows a similar trend to ME.
% Note that hyperparameters are not altered for EnKode at each sampling iteration, while hyperparameter optimization is used at each iteration for GP, which may contribute to this difference in performance, albeit small in general.
% This is an advantage of EnKode over the GP--it offers the flexibility of early termination for lower modeling times at the cost of higher errors, or the option to allow the model to further optimize over the data for lower errors.

% \noindent 
\vspace{-2mm}
\subsection{Bickley Jet}

The Bickley Jet is a complex flow, exhibiting a thin jet traversing from left to right and oscillating along the center line of the domain, inducing counter rotating vortices above and below this stream.

\begin{figure}[H]
    \centering
    \includegraphics[width=0.4785\textwidth]{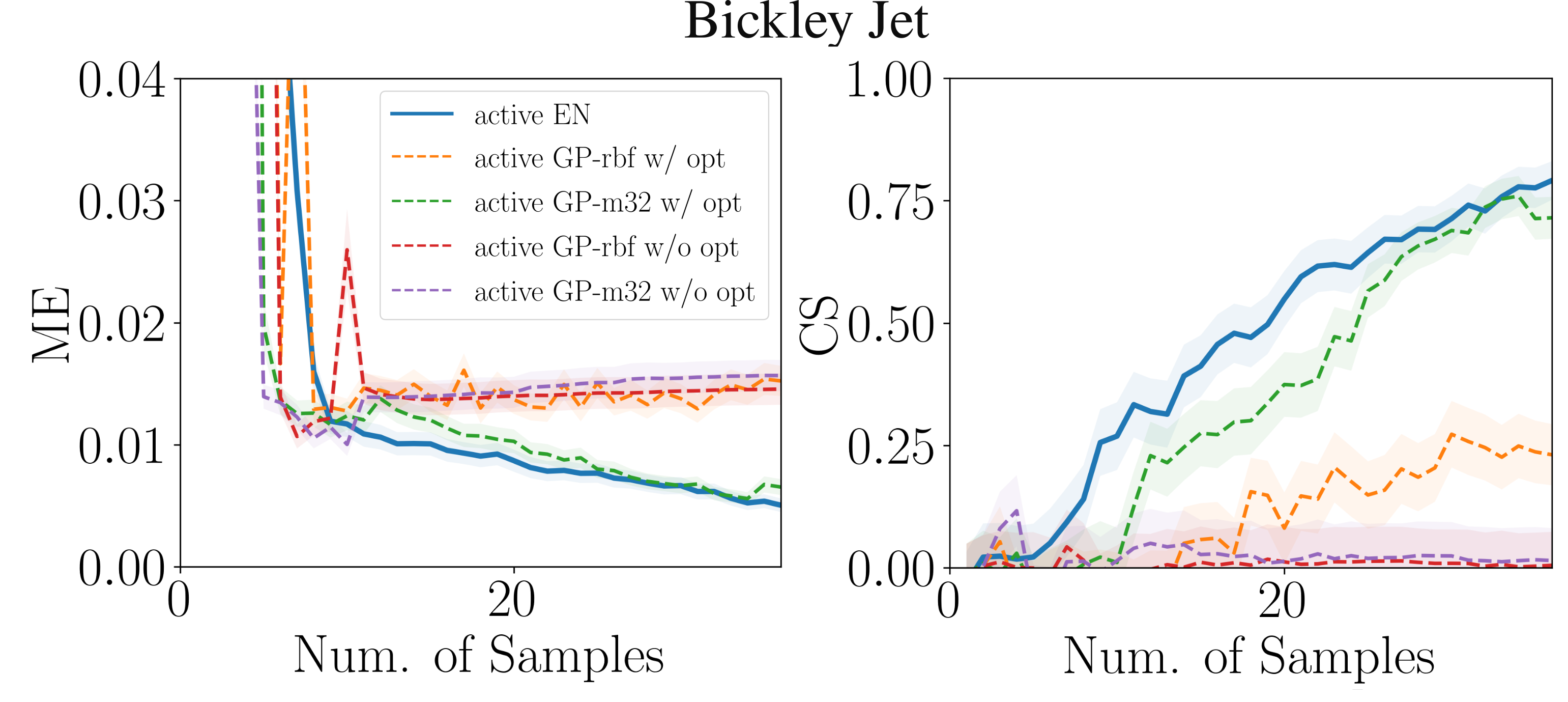}
    \caption{Errors when actively modeling Bickley Jet.
    ME values closer to $0$ are desired, while CS closer to $1$ are desired.
    For $N \geq 10$, ME is lower and CS is higher when actively sampling with EnKode.
    }
    % \Alice{Think about whether it makes more sense to show `sparser' sampling with less fancy looking plots or more dense and better looking plots, or both? Also, plot vortices instead of streamplots? Also, prohibiting from region with the cross makes most sense}
    \normalsize
\label{fig:metrics_bickley}
\end{figure}

This experiment demonstrates the effectiveness of combining Koopman theory and our uncertainty measure for active sampling.
In comparing Table \ref{tab:metrics_active} with Table \ref{tab:metrics_uniform}, we see that active sampling is better than uniform sampling for both EnKode and the GP. 
Furthermore, when actively sampling, Table \ref{tab:metrics_active} shows that CS is higher for EnKode in comparison to either GP, showing greater angular alignment of the estimated velocity field for any number of samples. ME for EnKode is lower or comparable to either GP. 
Active sampling spatial errors are shown in Fig. \ref{fig:flows_bickley} with $36$ samples.
There are regions with high errors for the GP that do not exist when actively sampling and modeling with EnKode.
Finally, trends in Fig. \ref{fig:metrics_bickley} show that EnKode metrics are better than both GP \textsc{w/ opt}.

% \vspace{-1mm}

\subsection{Lid-Driven Cavity with Obstacle}
To add complexity to the standard Lid-Driven Cavity flow, we embed an obstacle (cross shape) in the middle of the cavity. 
The bulk of the flow in this domain is a large clockwise rotating vortex. 
The obstacles, namely the cross and walls, give rise to smaller vortex structures---it can be challenging to capture features with % drastically 
different length scales. 

\begin{figure}[H]
    \centering    \includegraphics[width=0.485\textwidth]{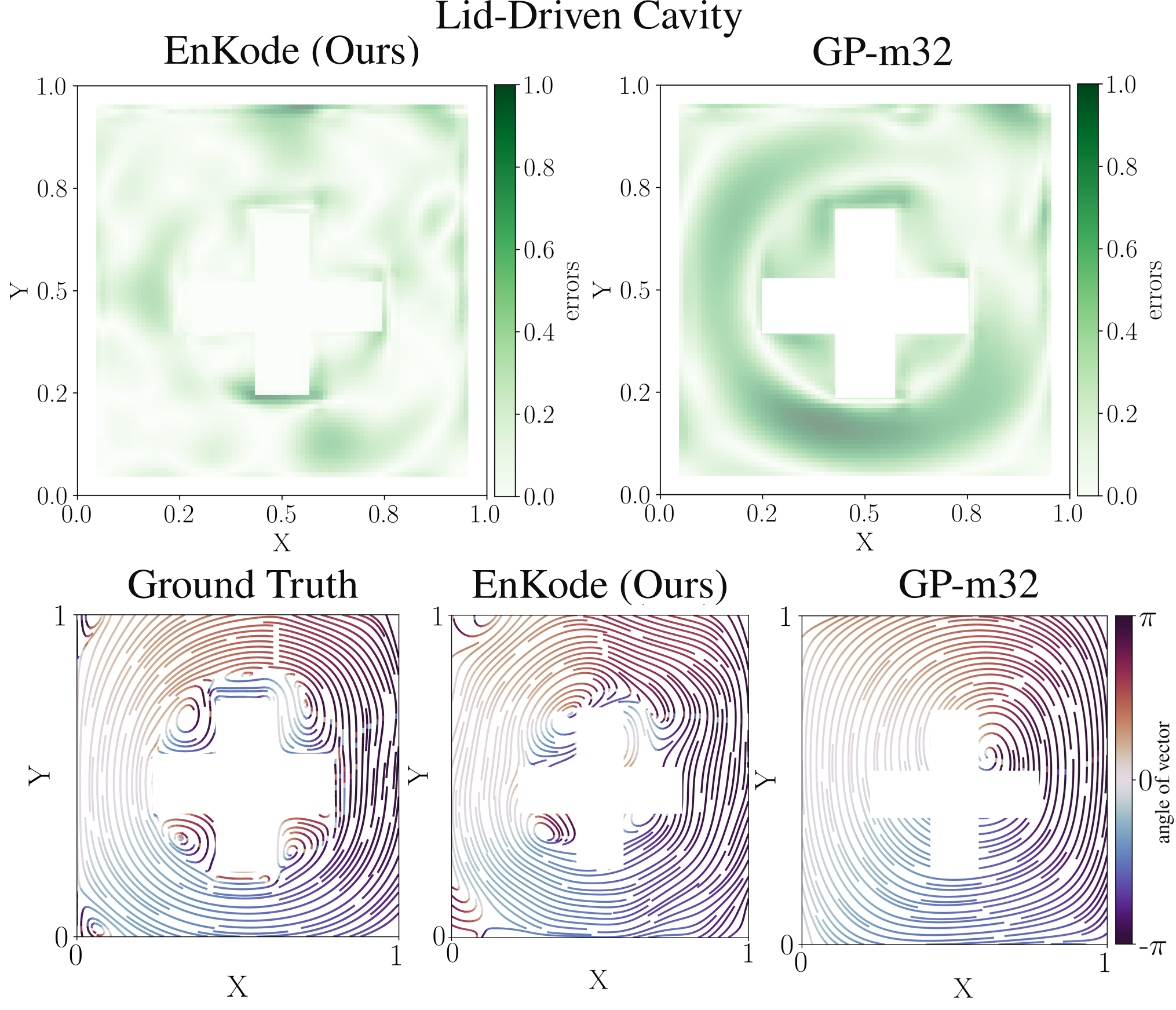}
    \caption{\textsc{Top.} Pixel-wise EPE for Lid-Driven Cavity with Obstacle. \textsc{Bottom.} Corresponding streamlines. When actively sampling, EnKode captures both the bulk flow dynamics and small vortices, while the GP is able to capture the bulk flow only.}
    \normalsize
\label{fig:flow_cavity}
\end{figure}

\begin{figure}[H]
    \centering    \includegraphics[width=0.4875\textwidth]{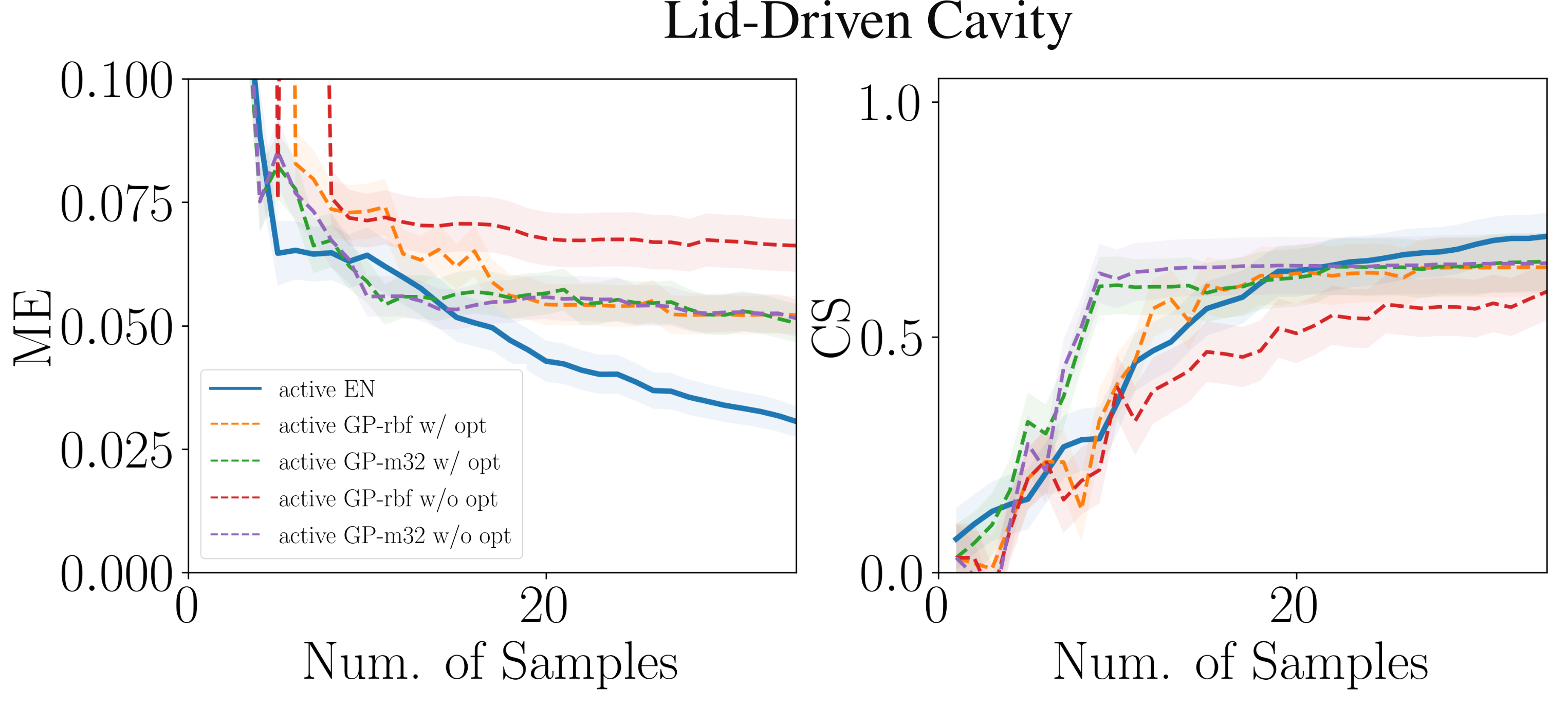}
    \caption{Errors for modeling the Lid-Driven Cavity flow with obstacle over $N \leq 36$ samples. 
    % ME values closer to $0$ and CS closer to $1$ are desired. 
    % With $36$ acquired samples, uniform sampling has lower errors, with GP-m32 and EnKode performing similarly. 
    If active sampling is used, EnKode performs better than the GP baseline. 
    }
    % \Alice{Think about whether it makes more sense to show `sparser' sampling with less fancy looking plots or more dense and better looking plots, or both? Also, plot vortices instead of streamplots? Also, prohibiting from region with the cross makes most sense}
    \normalsize
\label{fig:metrics_cavity}
\end{figure}

Unsurprisingly, uniform sampling in Table \ref{tab:metrics_uniform} is preferred over active sampling in Table \ref{tab:metrics_active}, as the flow structure is uniform across the majority of the domain.
When uniformly sampling, GP-m32 performs better than EnKode which performs better than GP-rbf.
However, Fig. \ref{fig:metrics_cavity} shows that when actively sampling, EnKode CS and ME performance exceeds either GP after $20$ samples have been collected, as Eq. \eqref{eqn:informativeness} guides the robot to sample around the embedded cross in Fig. \ref{fig:flow_cavity}. 
Fig. \ref{fig:flow_cavity} better illustrates how EnKode is able to capture the regions around the obstacle that exhibits smaller clockwise rotating vortices, while the GP cannot. 
In addition to showing that active sampling with EnKode performs better compared to the GP baseline, we also note that this is a successful demonstration of EnKode modeling obstacle-laden flow.
% EnKode outperforms the GP when performing active sensing, consistently achieving higher CS and lower ME. 
% , but for the GP, the active sampling performance is better with 
% dominates  % no longer using, since Ani recommended restricting to less samples to highlight EnKode performance.
% less data and uniform performance better with more. 
% If restricted to a limited sampling budget, EnKode with active sampling is preferred in this domain.

% Results in Table \ref{tab:metrics_active} also confirm that EnKode is better than the GP in actively modeling the Bickley Jet, whether modeled with the m-32 kernel or rbf, and for any number of samples. This is demonstrated by the greater CS and lower ME values.

% \subsection{Lid-Driven Cavity with Obstacle}
% To add complexity to the standard Lid-Driven Cavity flow, we embed an obstacle (cross shape) in the middle of the cavity. 
% The bulk of the flow in this domain is a large clockwise rotating vortex. 
% The obstacles, namely the cross and walls, give rise to smaller vortex structures---it can be challenging to capture features with drastically different length scales. 

% In Table \ref{tab:metrics}, we see that the errors for the GP may start off better than EnKode with $16$ samples, but do not improve with $36$.
\vspace{-3mm}
\subsection{Real Ocean Currents}
In comparison to the previous two flows, ocean currents exhibit less symmetry. A good model should be able to capture features with varying degrees of magnitude and vorticity, and not smooth out the features with smaller magnitudes.

\begin{figure}[H]
    \centering    \includegraphics[width=0.4875\textwidth]{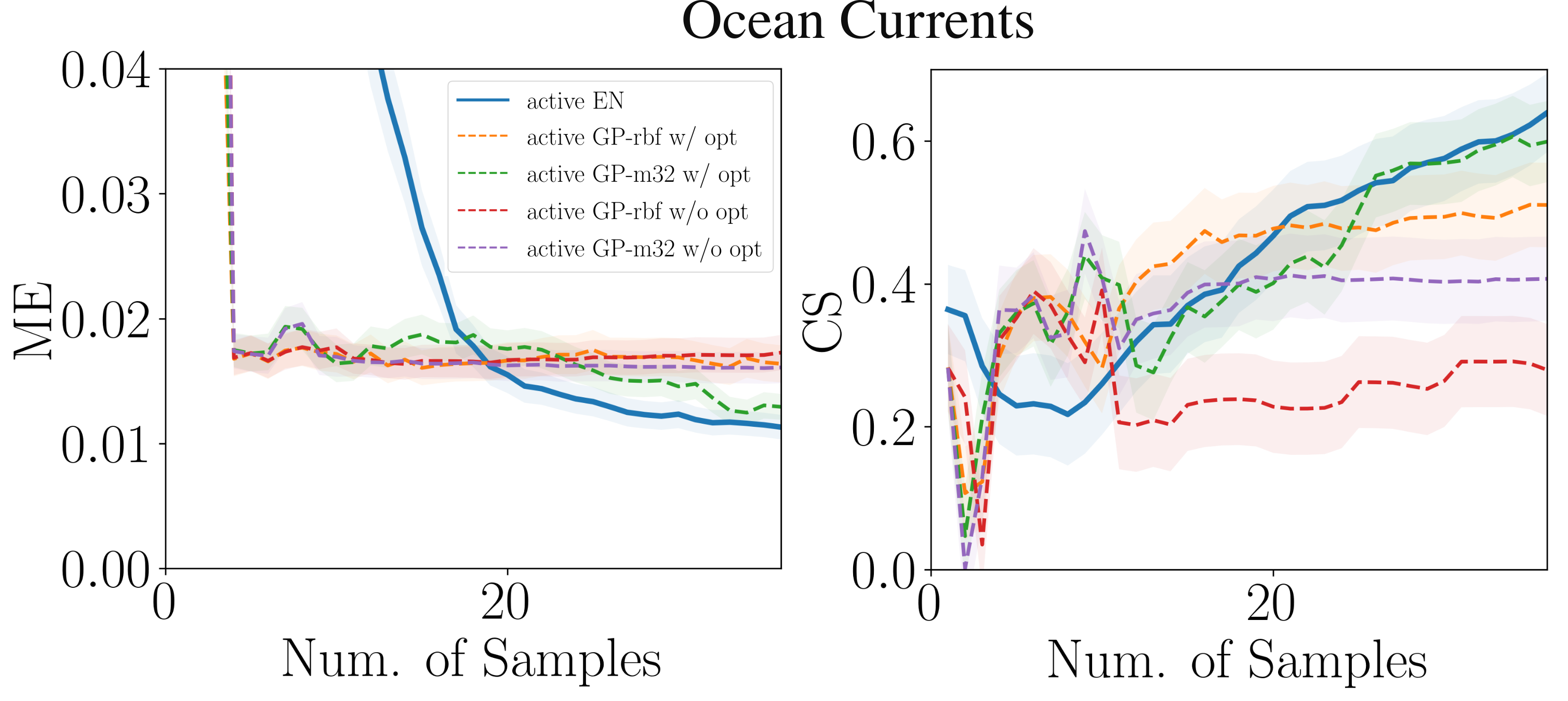}
    \caption{Errors for modeling ocean currents over $N \leq 36$ samples.
    % ME values closer to $0$ are desired, while CS closer to $1$ are desired. 
    Active sampling with EnKode modeling provides reliable performance after $20$ samples.
    GP performance occasionally exceeds EnKode, but the performance is less consistent, which is undesirable.}
    % \Alice{Think about whether it makes more sense to show `sparser' sampling with less fancy looking plots or more dense and better looking plots, or both? Also, plot vortices instead of streamplots? Also, prohibiting from region with the cross makes most sense}
    \normalsize
\label{fig:metrics_ocean}
\end{figure}

After acquiring $20$ samples selected via adaptive sampling, EnKode performs better than both GPs, as observed in Fig. \ref{fig:metrics_ocean}.
While active sampling with GP can occasionally outperform EnKode uniform and active sampling, the error fluctuations are undesirable, possibly capturing the GP sensitivity to training data and hyperparameter selection.
By the end of the active sensing task, $N=36$, active sampling is better than uniform sampling for both EnKode and GP, when comparing results across \ref{tab:metrics_active} and \ref{tab:metrics_uniform}.
For uniform sampling in \ref{tab:metrics_uniform}, the initial well-performing GP is likely because the grid of samples represent the general flow of the domain.
%With additional uniform samples, GP performance plateaus while EnKode increases.
For a more consistent performance when modeling ocean currents, EnKode should be adopted with the active sampling scheme.
Again, Table \ref{tab:metrics_active} further confirms that actively sampling with EnKode is effective. 
EnKode performs better than both GP models for both metrics when modeling with $N \geq 25$.

\begin{figure*} %[H]
    \centering
    \includegraphics[width=1\textwidth]{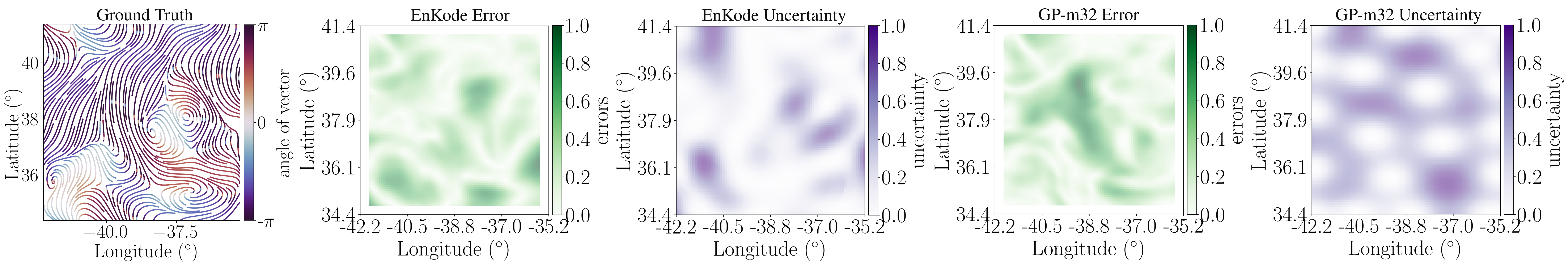}
    \caption{\textcolor{black}{Spatial plots of ground-truth flow, pixel-wise EPE, and uncertainty for EnKode and GP-m32, when modeling ocean currents. For EnKode, there is overlap in the spatial regions with high values of prediction error and uncertainty, indicating that the EnKode uncertainty estimation is meaningful for both defining the next best sampling location in an information gathering scenario, as well as serving as a suitable method for flow prediction uncertainty. On the other hand, the uncertainty values for the GP is low in sampled locations, and high elsewhere.}}
    % \Alice{Think about whether it makes more sense to show `sparser' sampling with less fancy looking plots or more dense and better looking plots, or both? Also, plot vortices instead of streamplots? Also, prohibiting from region with the cross makes most sense}
    \normalsize
\label{fig:error_vs_uncertainty}
\end{figure*}

Overall, these experiments demonstrate that the uncertainty metric we define for active sampling is effective, not only for simulated data, but also using real ocean model data.
Fig. \ref{fig:error_vs_uncertainty} shows a comparison between pixel-wise errors and uncertainty after sampling at $20$ locations, demonstrating the effectiveness of the informativeness measure in \eqref{eqn:informativeness}.
Furthermore, the results highlight that the EnKode uncertainty estimation is more meaningful as it more closely represents the prediction error compared to that of the GP baseline.

\vspace{-2mm}
\subsection{Sensitivity Analyses}
\textbf{Information Weighting, $\beta$.}
Table \ref{tab:sensitivity_beta} shows the effect of the weighting in Eq. \eqref{eqn:informative_weighting}. 
% The plot shows a cumulative sum of errors after $50$ samples have been acquired. 
A small $\beta$ value places more importance on sampling in regions with high magnitude variance, and a large $\beta$ value encourages sampling in regions with high angular variance. 
We observe that lower ME errors are obtained when $\beta$ is more heavily weighted towards one type of variance. 
The best CS value is obtained when $\beta=10
$, while the best ME value is obtained when $\beta=0.1$. 
However, in general, there is no significant trend when $\beta$ is varied.

\textbf{Number of Fourier Features, $\nu$.} 
Given a fixed number of optimization iterations, there is an ideal number of Fourier features used to model the ocean flow as shown in Table \ref{tab:sensitivity_beta}.
In this experiment, we optimize over Eq. \eqref{eqn:loss} with a fixed number of iterations, and an $8 \times 8$ uniform grid of samples. 
\vspace{-2mm}

\vspace{-2mm}

\vspace{-2mm}

\vspace{1mm}
\noindent 
We evaluate the effect of increasing the number of Fourier features used to define the observables.
We observe that for the number of iterations selected, $2^5$ Fourier features is ideal for the lowest ME error. 
The CS is best with $2^6$ features. 
The increased errors with few features is likely from a lack of expressivity; while errors from many features is likely due to a lack of estimation convergence or overfitting.

\textbf{Ensemble Size, M.} 
Table. \ref{tab:sensitivity_beta} shows the effect of the number of ensemble members. 
To compute metrics in the sensitivity analyses, we accumulate the sum of errors for $N = 50$ samples. 
In general, the more members, the lower the ME. 
The CS is marginally greater with the largest number of ensemble members.
There is no significant trend for CS versus $M$.
We note that the performance across $10$ experiments with $5$ ensemble members is much more varied, and is a possible reason why the result for $5$ ensemble members does not fit the same trend as the results for other ensemble sizes.

\begin{table}[htbp]
    % \fontsize{6}{4} 
    \vspace{8mm}
    \centering 
    \caption{\textsc{Effect of $\beta$, $\nu$ and $M$ for active sampling.}}
    % Errors and variance across $10$ are shown.}} 
    \NewDocumentCommand{\B}{}{\fontseries{b}\selectfont}
    \centering
    \resizebox{\linewidth}{!}{%
    \begin{tabular}{
        @{}
        l
        c c c c c 
        @{}
    }
        \toprule
        $\beta$ & {0.1} & {0.5} & {1} & {2} & {10} \\
        \midrule
        CS $(\uparrow)$ &2.326 (1.395)& 2.272 (1.389)	&2.362 (1.426)	&2.358 (1.417)	& \textbf{2.385} (1.282) \\
        ME $(\downarrow)$ & \textbf{1.023} (0.278)&	1.172 (0.268)&	1.133 (0.266)&	1.044 (0.274)	& 1.051 (0.319) \\ 
        \bottomrule
    \end{tabular}
    }
    
     \resizebox{\linewidth}{!}{%
    \begin{tabular}{
        @{}
        l
        c c c c c 
        @{}
    }
        \toprule
        $\nu$ & $2^3$ & $2^4$ & $2^5$ & $2^6$ & $2^7$ \\
        \midrule
        CS $(\uparrow)$& 0.711 (0.044)	& 0.836 (0.018) & 0.876 (0.009) & \textbf{0.883} (0.005) & 0.854 (0.014) \\
        ME $(\downarrow)$	& 0.029 (3E-3) & 0.018 (9E-4) & \textbf{0.015} (6E-4) & 0.016 (6E-4) &	0.023 (1E-3) \\
        \bottomrule
    \end{tabular}
    }
    
    \resizebox{\linewidth}{!}{%
    \begin{tabular}{
        @{}
        l
        c c c c c 
        @{}
    }
        \toprule
        $M$ & {5} & {10} & {15} & {20} & {25} \\
        \midrule
        CS $(\uparrow)$	&2.338 (1.347)	&2.333 (1.444)	&2.366 (1.525)	&2.330 (1.442)	& \textbf{2.447} (1.400) \\
        ME $(\downarrow)$&1.105 (0.272)	&1.179 (0.253)	&1.162 (0.251)	&1.088 (1.174)	& \textbf{1.019} (0.258) \\
        \bottomrule
    \end{tabular}
    }
    \label{tab:sensitivity_beta}
    % \vspace{-4mm}
\end{table}
\vspace{-4mm}

\section{Conclusions}
In this letter, we propose EnKode, a framework that provides flow estimation and uncertainty quantification for robotic adaptive sampling. 
We demonstrate the efficacy of EnKode in three different flow environments.
Through these experiments, we show that KOT allows for high quality predictions in unsampled flow regions.
\textcolor{black}{This, along with its other benefits, including linear analysis, control, and physical interpretability, make methods grounded in operator theory favorable over dynamics-agnostic flow models.}
Results also show that the EnKode informativeness criterion and uncertainty quantification with ensemble methods is \textcolor{black}{effective} for determining the next best sampling location, as it gives rise to a reduction in the overall modeling error for three different metrics (end-to-end point error, magnitude error, and cosine similarity).
When comparing sampling strategies and modeling methods, we find that for more complex flows---those with higher vortex structure like the Bickley Jet---active sampling is preferred over uniform sampling, and EnKode modeling is preferred over GP modeling.

This work provides an opportunity for many interesting directions for future work.
\textcolor{black}{This work could be extended to actively learning the dynamics of higher dimensional systems, where larger amounts of data would likely be required to learn the system dynamics.}
% The active learning strategy could be further improved by estimating the expected entropy reduction from sampling in a certain spatial location. 
In addition, in this work, the next best sampling location is intentionally point-based to assess the modeling and uncertainty estimation of EnKode.
In future work, planning can be trajectory-based and also subject to travel budget and robot kinodynamic constraints.
Furthermore, our approach can be extended to modeling time-varying flows, \textcolor{black}{known to be periodic, as in \cite{mezic2016koopman}}, and planning with multi-robot sytems.

% \begin{itemize}
%     \item good information theoretic measure for active sampling - as shown in active sampling performance and the similarity in the spatial locations with errors vs uncertainty
%     \item active sampling is better than uniform sampling as a result 
%     \item EnKode modeling is better than GP for more complex flows
%     \item future work: extension to trajectory based planning with a sampling budget 
%     \item time-varying Koopman operator 
% \end{itemize}

% \addtolength{\textheight}{-12cm}   % This command serves to balance the column lengths
                                  % on the last page of the document manually. It shortens
                                  % the textheight of the last page by a suitable amount.
                                  % This command does not take effect until the next page
                                  % so it should come on the page before the last. Make
                                  % sure that you do not shorten the textheight too much.

%%%%%%%%%%%%%%%%%%%%%%%%%%%%%%%%%%%%%%%%%%%%%%%%%%%%%%%%%%%%%%%%%%%%%%%%%%%%%%%%

%%%%%%%%%%%%%%%%%%%%%%%%%%%%%%%%%%%%%%%%%%%%%%%%%%%%%%%%%%%%%%%%%%%%%%%%%%%%%%%%

%%%%%%%%%%%%%%%%%%%%%%%%%%%%%%%%%%%%%%%%%%%%%%%%%%%%%%%%%%%%%%%%%%%%%%%%%%%%%%%%
\vspace{-4mm}
\section*{Acknowledgements}
We greatly acknowledge the support of NSF IUCRC 1939132, NSF IIS 1910308, and the University of Pennsylvania’s University Research Foundation Award. We would also like to thank Spencer Folk for data generation of the Lid-driven Cavity flow.

%%%%%%%%%%%%%%%%%%%%%%%%%%%%%%%%%%%%%%%%%%%%%%%%%%%%%%%%%%%%%%%%%%%%%%%%%%%%%%%%
%\section*{References}
\vspace{-2mm}
\bibliographystyle{IEEEtran} % ieeetr}
\bibliography{citations}

% Generated by IEEEtran.bst, version: 1.14 (2015/08/26)
\begin{thebibliography}{10}
\providecommand{\url}[1]{#1}
\csname url@samestyle\endcsname
\providecommand{\newblock}{\relax}
\providecommand{\bibinfo}[2]{#2}
\providecommand{\BIBentrySTDinterwordspacing}{\spaceskip=0pt\relax}
\providecommand{\BIBentryALTinterwordstretchfactor}{4}
\providecommand{\BIBentryALTinterwordspacing}{\spaceskip=\fontdimen2\font plus
\BIBentryALTinterwordstretchfactor\fontdimen3\font minus \fontdimen4\font\relax}
\providecommand{\BIBforeignlanguage}[2]{{%
\expandafter\ifx\csname l@#1\endcsname\relax
\typeout{** WARNING: IEEEtran.bst: No hyphenation pattern has been}%
\typeout{** loaded for the language `#1'. Using the pattern for}%
\typeout{** the default language instead.}%
\else
\language=\csname l@#1\endcsname
\fi
#2}}
\providecommand{\BIBdecl}{\relax}
\BIBdecl

\bibitem{dunbabin2012robots}
M.~Dunbabin and L.~Marques, ``Robots for environmental monitoring: Significant advancements and applications,'' \emph{IEEE Robotics \& Automation Magazine}, vol.~19, no.~1, pp. 24--39, 2012.

\bibitem{to2021estimation}
K.~C. To, F.~H. Kong, K.~M.~B. Lee, C.~Yoo, S.~Anstee, and R.~Fitch, ``Estimation of spatially-correlated ocean currents from ensemble forecasts and online measurements,'' in \emph{2021 IEEE International Conference on Robotics and Automation (ICRA)}.\hskip 1em plus 0.5em minus 0.4em\relax IEEE, 2021, pp. 2301--2307.

\bibitem{hansen2018coverage}
J.~Hansen and G.~Dudek, ``Coverage optimization with non-actuated, floating mobile sensors using iterative trajectory planning in marine flow fields,'' in \emph{2018 IEEE/RSJ International Conference on Intelligent Robots and Systems (IROS)}.\hskip 1em plus 0.5em minus 0.4em\relax IEEE, 2018, pp. 1906--1912.

\bibitem{khodayi2023physics}
R.~Khodayi-mehr, P.~Jian, and M.~M. Zavlanos, ``Physics-guided active learning of environmental flow fields,'' in \emph{Learning for Dynamics and Control Conference}.\hskip 1em plus 0.5em minus 0.4em\relax PMLR, 2023, pp. 928--940.

\bibitem{chang2017motion}
D.~Chang, W.~Wu, C.~R. Edwards, and F.~Zhang, ``Motion tomography: Mapping flow fields using autonomous underwater vehicles,'' \emph{The International Journal of Robotics Research}, vol.~36, no.~3, pp. 320--336, 2017.

\bibitem{michini2014robotic}
M.~Michini, M.~A. Hsieh, E.~Forgoston, and I.~B. Schwartz, ``Robotic tracking of coherent structures in flows,'' \emph{IEEE Transactions on Robotics}, vol.~30, no.~3, pp. 593--603, 2014.

\bibitem{molchanov2015active}
A.~Molchanov, A.~Breitenmoser, and G.~S. Sukhatme, ``Active drifters: Towards a practical multi-robot system for ocean monitoring,'' in \emph{2015 IEEE International Conference on Robotics and Automation (ICRA)}.\hskip 1em plus 0.5em minus 0.4em\relax IEEE, 2015, pp. 545--552.

\bibitem{hollinger2016learning}
G.~A. Hollinger, A.~A. Pereira, J.~Binney, T.~Somers, and G.~S. Sukhatme, ``Learning uncertainty in ocean current predictions for safe and reliable navigation of underwater vehicles,'' \emph{Journal of Field Robotics}, vol.~33, no.~1, pp. 47--66, 2016.

\bibitem{salam2019adaptive}
T.~Salam and M.~A. Hsieh, ``Adaptive sampling and reduced-order modeling of dynamic processes by robot teams,'' \emph{IEEE Robotics and Automation Letters}, vol.~4, no.~2, pp. 477--484, 2019.

\bibitem{folk2024learning}
S.~Folk, J.~Melton, B.~W. Margolis, M.~Yim, and V.~Kumar, ``Learning local urban wind flow fields from range sensing,'' \emph{IEEE Robotics and Automation Letters}, 2024.

\bibitem{karniadakis2021physics}
G.~E. Karniadakis, I.~G. Kevrekidis, L.~Lu, P.~Perdikaris, S.~Wang, and L.~Yang, ``Physics-informed machine learning,'' \emph{Nature Reviews Physics}, vol.~3, no.~6, pp. 422--440, 2021.

\bibitem{koopman1931hamiltonian}
B.~O. Koopman, ``Hamiltonian systems and transformation in hilbert space,'' \emph{Proceedings of the National Academy of Sciences}, vol.~17, no.~5, pp. 315--318, 1931.

\bibitem{mezic2013analysis}
I.~Mezi{\'c}, ``Analysis of fluid flows via spectral properties of the koopman operator,'' \emph{Annual review of fluid mechanics}, vol.~45, pp. 357--378, 2013.

\bibitem{budivsic2012applied}
M.~Budi{\v{s}}i{\'c}, R.~Mohr, and I.~Mezi{\'c}, ``Applied koopmanism,'' \emph{Chaos: An Interdisciplinary Journal of Nonlinear Science}, vol.~22, no.~4, 2012.

\bibitem{bevanda2021koopman}
P.~Bevanda, S.~Sosnowski, and S.~Hirche, ``Koopman operator dynamical models: Learning, analysis and control,'' \emph{Annual Reviews in Control}, vol.~52, pp. 197--212, 2021.

\bibitem{otto2021koopman}
S.~E. Otto and C.~W. Rowley, ``Koopman operators for estimation and control of dynamical systems,'' \emph{Annual Review of Control, Robotics, and Autonomous Systems}, vol.~4, pp. 59--87, 2021.

\bibitem{williams2015data}
M.~O. Williams, I.~G. Kevrekidis, and C.~W. Rowley, ``A data--driven approximation of the koopman operator: Extending dynamic mode decomposition,'' \emph{Journal of Nonlinear Science}, vol.~25, pp. 1307--1346, 2015.

\bibitem{tu2013dynamic}
J.~H. Tu, ``Dynamic mode decomposition: Theory and applications,'' Ph.D. dissertation, Princeton University, 2013.

\bibitem{klus2020eigendecompositions}
S.~Klus, I.~Schuster, and K.~Muandet, ``Eigendecompositions of transfer operators in reproducing kernel hilbert spaces,'' \emph{Journal of Nonlinear Science}, vol.~30, pp. 283--315, 2020.

\bibitem{lusch2018deep}
B.~Lusch, J.~N. Kutz, and S.~L. Brunton, ``Deep learning for universal linear embeddings of nonlinear dynamics,'' \emph{Nature communications}, vol.~9, no.~1, p. 4950, 2018.

\bibitem{leask2021modal}
S.~Leask, V.~McDonell, and S.~Samuelsen, ``Modal extraction of spatiotemporal atomization data using a deep convolutional koopman network,'' \emph{Physics of Fluids}, vol.~33, no.~3, 2021.

\bibitem{schmid2022dynamic}
P.~J. Schmid, ``Dynamic mode decomposition and its variants,'' \emph{Annual Review of Fluid Mechanics}, vol.~54, pp. 225--254, 2022.

\bibitem{rasmussen2006gaussian}
C.~E. Rasmussen, C.~K. Williams \emph{et~al.}, \emph{Gaussian processes for machine learning}.\hskip 1em plus 0.5em minus 0.4em\relax Springer, 2006, vol.~1.

\bibitem{lee2019online}
K.~M.~B. Lee, C.~Yoo, B.~Hollings, S.~Anstee, S.~Huang, and R.~Fitch, ``Online estimation of ocean current from sparse gps data for underwater vehicles,'' in \emph{2019 International conference on robotics and automation (ICRA)}.\hskip 1em plus 0.5em minus 0.4em\relax IEEE, 2019, pp. 3443--3449.

\bibitem{berlinghieri2023gaussian}
R.~Berlinghieri, B.~L. Trippe, D.~R. Burt, R.~Giordano, K.~Srinivasan, T.~{\"O}zg{\"o}kmen, J.~Xia, and T.~Broderick, ``Gaussian processes at the helm (holtz): A more fluid model for ocean currents,'' \emph{arXiv preprint arXiv:2302.10364}, 2023.

\bibitem{gonccalves2019reconstruction}
R.~C. Gon{\c{c}}alves, M.~Iskandarani, T.~{\"O}zg{\"o}kmen, and W.~C. Thacker, ``Reconstruction of submesoscale velocity field from surface drifters,'' \emph{Journal of Physical Oceanography}, vol.~49, no.~4, pp. 941--958, 2019.

\bibitem{petrich2009planar}
J.~Petrich, C.~A. Woolsey, and D.~J. Stilwell, ``Planar flow model identification for improved navigation of small auvs,'' \emph{Ocean Engineering}, vol.~36, no.~1, pp. 119--131, 2009.

\bibitem{arbabi2018data}
H.~Arbabi, M.~Korda, and I.~Mezic, ``A data-driven koopman model predictive control framework for nonlinear flows,'' \emph{arXiv preprint arXiv:1804.05291}, 2018.

\bibitem{salam2022}
\BIBentryALTinterwordspacing
T.~Salam, A.~K. Li, and M.~A. Hsieh, ``Online estimation of the koopman operator using fourier features,'' in \emph{Learning for Dynamics and Control Conference, {L4DC} 2023}, ser. Proceedings of Machine Learning Research, vol. 211.\hskip 1em plus 0.5em minus 0.4em\relax {PMLR}, 2023, pp. 1271--1283. [Online]. Available: \url{https://proceedings.mlr.press/v211/salam23a.html}
\BIBentrySTDinterwordspacing

\bibitem{han2023utility}
Y.~Han, M.~Xie, Y.~Zhao, and H.~Ravichandar, ``On the utility of koopman operator theory in learning dexterous manipulation skills,'' in \emph{Conference on Robot Learning}.\hskip 1em plus 0.5em minus 0.4em\relax PMLR, 2023, pp. 106--126.

\bibitem{bruder2019modeling}
D.~Bruder, B.~Gillespie, C.~D. Remy, and R.~Vasudevan, ``Modeling and control of soft robots using the koopman operator and model predictive control,'' \emph{arXiv preprint arXiv:1902.02827}, 2019.

\bibitem{joglekar2023data}
A.~Joglekar, S.~Sutavani, C.~Samak, T.~Samak, K.~C. Kosaraju, J.~Smereka, D.~Gorsich, U.~Vaidya, and V.~Krovi, ``Data-driven modeling and experimental validation of autonomous vehicles using koopman operator,'' in \emph{2023 IEEE/RSJ International Conference on Intelligent Robots and Systems (IROS)}.\hskip 1em plus 0.5em minus 0.4em\relax IEEE, 2023.

\bibitem{folkestad2020episodic}
C.~Folkestad, D.~Pastor, and J.~W. Burdick, ``Episodic koopman learning of nonlinear robot dynamics with application to fast multirotor landing,'' in \emph{2020 IEEE International Conference on Robotics and Automation (ICRA)}.\hskip 1em plus 0.5em minus 0.4em\relax IEEE, 2020, pp. 9216--9222.

\bibitem{shi2021acd}
L.~Shi and K.~Karydis, ``Acd-edmd: Analytical construction for dictionaries of lifting functions in koopman operator-based nonlinear robotic systems,'' \emph{IEEE Robotics and Automation Letters}, vol.~7, no.~2, pp. 906--913, 2021.

\bibitem{abraham2019active}
I.~Abraham and T.~D. Murphey, ``Active learning of dynamics for data-driven control using koopman operators,'' \emph{IEEE Transactions on Robotics}, vol.~35, no.~5, pp. 1071--1083, 2019.

\bibitem{li2022towards}
A.~K. Li, Y.~Mao, S.~Manjanna, S.~Liu, J.~Dhanoa, B.~Mehta, V.~M. Edwards, F.~C. Ojeda, M.~A. Hsieh, M.~Le~Men \emph{et~al.}, ``Towards understanding underwater weather events in rivers using autonomous surface vehicles,'' in \emph{OCEANS 2022, Hampton Roads}.\hskip 1em plus 0.5em minus 0.4em\relax IEEE, 2022, pp. 1--8.

\bibitem{edwards2023collaborative}
V.~Edwards, T.~C. Silva, B.~Mehta, J.~Dhanoa, and M.~A. Hsieh, ``On collaborative robot teams for environmental monitoring: A macroscopic ensemble approach,'' in \emph{2023 IEEE/RSJ International Conference on Intelligent Robots and Systems (IROS)}.\hskip 1em plus 0.5em minus 0.4em\relax IEEE, 2023, pp. 11\,148--11\,153.

\bibitem{krause2007nonmyopic}
A.~Krause and C.~Guestrin, ``Nonmyopic active learning of gaussian processes: an exploration-exploitation approach,'' in \emph{Proceedings of the 24th international conference on Machine learning}, 2007, pp. 449--456.

\bibitem{taylor2021active}
A.~T. Taylor, T.~A. Berrueta, and T.~D. Murphey, ``Active learning in robotics: A review of control principles,'' \emph{Mechatronics}, vol.~77, p. 102576, 2021.

\bibitem{berrueta2020experimental}
T.~A. Berrueta, I.~Abraham, and T.~Murphey, ``Experimental applications of the koopman operator in active learning for control,'' \emph{The Koopman Operator in Systems and Control: Concepts, Methodologies, and Applications}, pp. 421--450, 2020.

\bibitem{settles2009active}
B.~Settles, ``Active learning literature survey,'' 2009.

\bibitem{seung1992query}
H.~S. Seung, M.~Opper, and H.~Sompolinsky, ``Query by committee,'' in \emph{Proceedings of the fifth annual workshop on Computational learning theory}, 1992, pp. 287--294.

\bibitem{pathak2019self}
D.~Pathak, D.~Gandhi, and A.~Gupta, ``Self-supervised exploration via disagreement,'' in \emph{International conference on machine learning}.\hskip 1em plus 0.5em minus 0.4em\relax PMLR, 2019, pp. 5062--5071.

\bibitem{georgakis2022learning}
\BIBentryALTinterwordspacing
G.~Georgakis, B.~Bucher, K.~Schmeckpeper, S.~Singh, and K.~Daniilidis, ``Learning to map for active semantic goal navigation,'' in \emph{International Conference on Learning Representations}, 2022. [Online]. Available: \url{https://openreview.net/forum?id=swrMQttr6wN}
\BIBentrySTDinterwordspacing

\bibitem{lakshminarayanan2017simple}
B.~Lakshminarayanan, A.~Pritzel, and C.~Blundell, ``Simple and scalable predictive uncertainty estimation using deep ensembles,'' \emph{Advances in neural information processing systems}, vol.~30, 2017.

\bibitem{brunton2021modern}
S.~L. Brunton, M.~Budi{\v{s}}i{\'c}, E.~Kaiser, and J.~N. Kutz, ``Modern koopman theory for dynamical systems,'' \emph{arXiv preprint arXiv:2102.12086}, 2021.

\bibitem{yeung2019learning}
E.~Yeung, S.~Kundu, and N.~Hodas, ``Learning deep neural network representations for koopman operators of nonlinear dynamical systems,'' in \emph{2019 American Control Conference (ACC)}.\hskip 1em plus 0.5em minus 0.4em\relax IEEE, 2019, pp. 4832--4839.

\bibitem{li2017extended}
Q.~Li, F.~Dietrich, E.~M. Bollt, and I.~G. Kevrekidis, ``Extended dynamic mode decomposition with dictionary learning: A data-driven adaptive spectral decomposition of the koopman operator,'' \emph{Chaos: An Interdisciplinary Journal of Nonlinear Science}, vol.~27, no.~10, 2017.

\bibitem{mauroy2019koopman}
A.~Mauroy and J.~Goncalves, ``Koopman-based lifting techniques for nonlinear systems identification,'' \emph{IEEE Transactions on Automatic Control}, vol.~65, no.~6, pp. 2550--2565, 2019.

\bibitem{shi2021cooperative}
L.~Shi, R.~Zheng, S.~Zhang, and M.~Liu, ``Cooperative estimation to reconstruct the parametric flow field using multiple auvs,'' \emph{IEEE Transactions on Instrumentation and Measurement}, vol.~70, pp. 1--10, 2021.

\bibitem{fisher1995statistical}
N.~I. Fisher, \emph{Statistical analysis of circular data}.\hskip 1em plus 0.5em minus 0.4em\relax cambridge university press, 1995.

\bibitem{simons2022erddap}
\BIBentryALTinterwordspacing
R.~Simons and C.~John, ``Erddap,'' 2022, monterey, CA: NOAA/NMFS/SWFSC/ERD. [Online]. Available: \url{https://coastwatch.pfeg.noaa.gov/erddap}
\BIBentrySTDinterwordspacing

\bibitem{otte1994optical}
M.~Otte and H.~H. Nagel, ``Optical flow estimation: advances and comparisons,'' in \emph{Computer Vision—ECCV'94: Third European Conference on Computer Vision Stockholm, Sweden, May 2--6, 1994 Proceedings, Volume I 3}.\hskip 1em plus 0.5em minus 0.4em\relax Springer, 1994, pp. 49--60.

\bibitem{gpy2014}
{GPy}, ``{GPy}: A gaussian process framework in python,'' \url{http://github.com/SheffieldML/GPy}, since 2012.

\bibitem{verner2024efficient}
M.-P. Verner~Christiansen, N.~R{\o}nne, and B.~Hammer, ``Efficient ensemble uncertainty estimation in gaussian processes regression,'' \emph{arXiv e-prints}, pp. arXiv--2407, 2024.

\bibitem{mezic2016koopman}
I.~Mezic and A.~Surana, ``Koopman mode decomposition for periodic/quasi-periodic time dependence,'' \emph{IFAC-PapersOnLine}, vol.~49, no.~18, pp. 690--697, 2016.

\end{thebibliography}
\end{document}